\documentclass[10pt,twocolumn,letterpaper]{article}

\usepackage{cvpr}
\usepackage{times}
\usepackage{epsfig}
\usepackage{graphicx}
\usepackage{amsmath}
\usepackage{amssymb}

\usepackage[utf8]{inputenc} 
\usepackage[T1]{fontenc}    
\usepackage{url}            
\usepackage{booktabs}       
\usepackage{amsfonts}       
\usepackage{nicefrac}       
\usepackage{microtype}      
\usepackage{amsmath,amssymb}
\usepackage{color}
\usepackage{array}
\usepackage{multirow}
\usepackage{booktabs}
\usepackage{subfigure}
\usepackage{amsthm}
\usepackage{wrapfig,lipsum,booktabs}
\usepackage{xcolor}

\usepackage[pagebackref=true,breaklinks=true,letterpaper=true,colorlinks,bookmarks=false]{hyperref}

\cvprfinalcopy 

\begin{document}

\title{DSRGAN: Explicitly Learning Disentangled Representation of  Underlying Structure and Rendering for Image Generation without Tuple Supervision}
\author{Guang-Yuan Hao$^{1}$ 
\footnotemark[1], Hong-Xing Yu$^{1,4}$ \footnotemark[1], Wei-Shi Zheng$^{1,2,3}$
\footnotemark[2]
\\
$^1$ School of Data and Computer Science, Sun Yat-sen University, China \\
$^2$ Key Laboratory of Machine Intelligence and Advanced Computing, Ministry of Education, China \\
$^3$ Collaborative Innovation Center of High Performance Computing, NUDT, China
\\
$^4$ Guangdong Key Laboratory of Big Data Analysis and Processing, Guangzhou, China
\\
guangyuanhao@outlook.com,
xKoven@gmail.com,
wszheng@ieee.org}

\maketitle
\renewcommand{\thefootnote}{\fnsymbol{footnote}}
\footnotetext[1]{Equal contribution}
\footnotetext[2]{Corresponding author}
\begin{abstract}
We focus on explicitly learning disentangled representation for natural image generation, where the underlying spatial structure and the rendering on the structure can be
independently controlled respectively, yet using no tuple supervision. The setting is significant since tuple supervision is costly and sometimes even unavailable.
However, the task is highly unconstrained and thus ill-posed.
To address this problem, we propose to introduce an auxiliary domain
which shares a common underlying-structure space with the target domain, and we make a partially shared latent space assumption.
The key idea is to encourage the partially shared latent variable
to represent the similar underlying spatial structures in both domains,
while the two domain-specific latent variables will be unavoidably arranged to present renderings of two domains respectively. This is achieved by designing two parallel generative networks with a common Progressive Rendering Architecture (\textbf{PRA}), which constrains both generative networks' behaviors to model shared underlying structure and to model spatially dependent relation between rendering and underlying structure. Thus, we propose DSRGAN (\textbf{GANs} for \textbf{D}isentangling Underlying \textbf{S}tructure and \textbf{R}endering) to instantiate our method. We also propose a quantitative criterion (the Normalized Disentanglability)  to quantify disentanglability.
Comparison to the state-of-the-art methods shows that DSRGAN can significantly outperform them in disentanglability.
\end{abstract}

\section{Introduction}
\begin{figure}[t!]
\small
\centering
\includegraphics[width=1.0\linewidth]{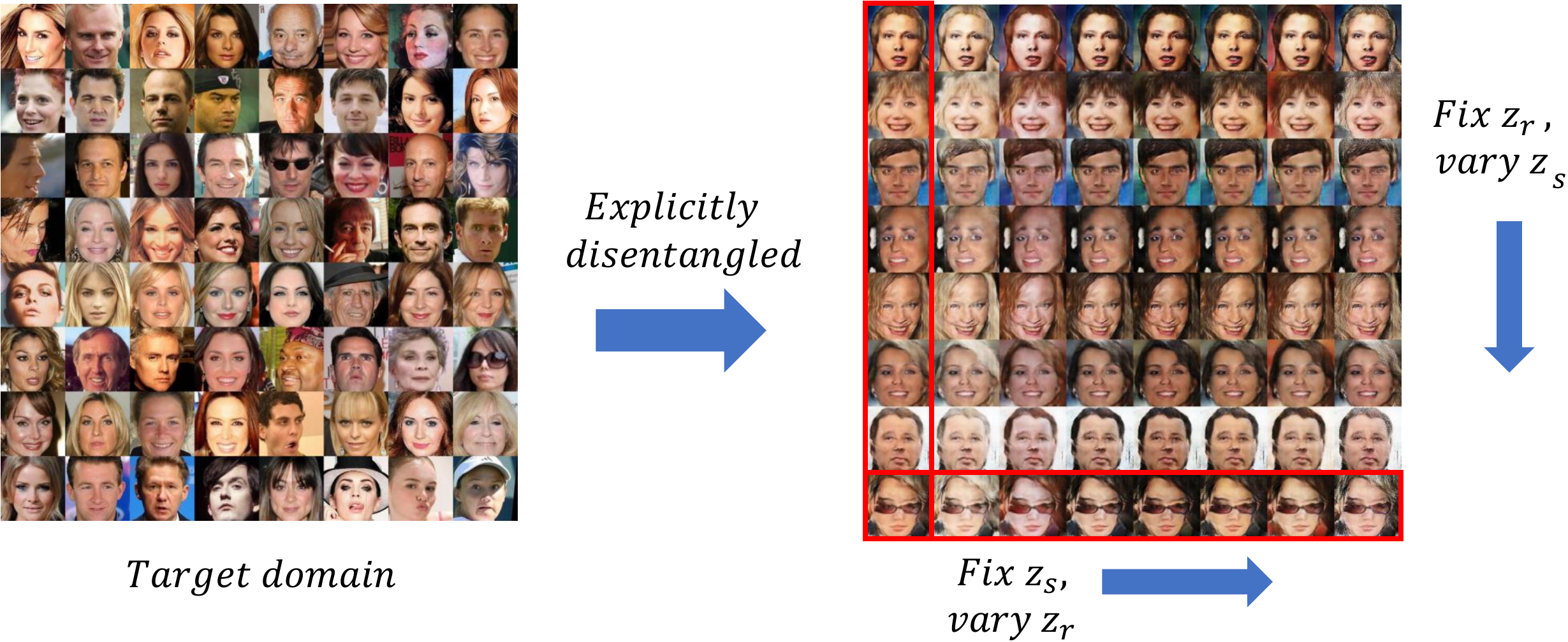}
\vspace{-0.45cm}
\caption{
Illustration of our aim. Without tuple supervision, we expect target domain to be explicitly disentangled into underlying spatial structure and rendering which are represented by  two latent variable $z_s$ and $z_r$ respectively.  (Best viewed in color.)}
\label{fig:intro}
\vspace{-0.55cm}
\end{figure}

In natural image generation, a suitable generative process might consist of two stages,
first of which is to generate an underlying spatial structure of the image, e.g. the shapes of a pair of sneakers or a facial skeleton of a human face.
The second stage involves rendering the underlying spatial structure,
e.g. the color style of the sneakers or the skin and hair of the face, to obtain a concrete image.
It is often desirable to have these two generative stages disentangled.
For example, a shoes designer can draw inspiration from various generated shoes images.
When she finds a eureka shape in some images, she may probably want to vary the color style for further exploration,
and the reverse procedure can be equally attractive to her. Another promising application is for data augmentation in deep learning \cite{pose} to improve robustness for face recognition, e.g., disentangling face ID and hair color or whether wearing eyeglasses can be applied to improve robustness of color-invariant or eyeglasses-invariant face recognition, so that when a person often changes hair color or wears kinds of eyeglasses, recognition system can stably recognize the person.

To explicitly disentangle the underlying spatial structure and rendering in the generative process,
we have to collect substantial label information across \emph{each} of these factors/dimensions, respectively\cite{SDGAN}.
This is highly costly or even impossible if the collected training data does not have such nature,
e.g., when we aim to disentangle hair color (regarded as rendering) from human identity (regarded as underlying spatial structure) in face images, we may need tuple supervision, i.e., collecting and annotating images of one person with several hair colors or images of several persons with the same hair color. Hence, in this work, we are interested in learning disentangled presentation in the generative process without any tuple supervision. Since this problem setting is different from previous work 
\cite{SDGAN}
, we refer to this problem as disentangled image generation without tuple supervision. As illustrated in Figure \ref{fig:intro},  we focus on learning a generative function $G_t(z_s, z_r)$ from a target domain where $z_s$ and $z_r$ are expected to
fully and only control the underlying spatial structure and the rendering of the generated images, respectively. Specially, as faces showed in red horizontal and vertical boxes in Figure \ref{fig:intro}, when we vary $z_r$ and fix $z_s$, faces possess kinds of hair colors with the same face ID and vice versa.  Although the setting without tuple supervision is general and useful, this task is very challenging and ill-posed since it is highly unconstrained.
Therefore, directly disentangling representations without explicit guidance will
lead to uncontrollable, ambiguous results \cite{betaVAE,infoGAN}.

To address this problem, we propose to introduce an auxiliary domain which shares common underlying-structure space with the target domain and has its specific rendering space.  We also assume two domains share partial latent variables \cite{2018_Arxiv_multimodal}.  The critical idea is to make the partially shared prior $z_s$ represent common factors of two domains, i.e., shared underlying structure and thus the rest of input priors $z_r$ ( i.e., domain-specific priors) will be unavoidably arranged to represent rendering.  As attributes of images contain no other factors except underlying structure and rendering, when underlying structure and rendering are represented by partially shared latent prior $z_s$ and domain-specific prior $z_r$ respectively,  we actually explicitly disentangle the target domain into its only two factors, i.e., underlying structure and rendering, which are represented by priors $z_s$ and $z_r$ successfully.

Specially, we propose DSRGAN (GANs for Disentangling Underlying Structure and Rendering) to instantiate our method
with a proposed Progressive Rendering Architecture (PRA), which enforces the partially shared latent prior to represent underlying structure and models the spatially dependent relation between rendering and underlying structure. We summarize our contributions as follows:

(1) We propose to introduce an auxiliary domain to provide explicit guidance to learn disentangled factors of interest without tuple supervision.

(2) We propose a novel framework DSRGAN to explicitly disentangle underlying structure and rendering with a proposed Progressive Rendering Architecture.

(3) We evaluate DSRGAN in several disentangled image generation tasks.
Since a proper quantitative measure of disentanglability is missing, we propose the Normalized Disentanglability to quantify disentanglability, which jointly formulates diversity and independence in the disentangling task.
Experimental results show that DSRGAN  significantly outperforms the state-of-the-art methods in disentanglability.

\section{Related Work}
\noindent
\textbf{Disentangled representation learning}.
Our work is related to disentangled representation learning.
One stream of previous works focuses on learning disentangled presentation in an unsupervised manner \cite{infoGAN,betaVAE,2018_ICLR_DIPVAE}. Those methods aim to make each dimension of the input prior represent one of unknown factors as fully as possible. Therefore, factors that they disentangle are random and ambiguous and thus those models do not always disentangle factors people care. Our method is different since we design a specific architecture which focuses on explicitly disentangling underlying structure and rendering. Hence, our model can explicitly specify and control the factors of interest.

Another stream is to disentangle some specific factor of interest from other irrelevant factors, typically requiring tuple supervision \cite{SDGAN,2017_NIPS_bicyclegan}
and input condition for image-to-image translation\cite{2018_Arxiv_multimodal,2017_NIPS_bicyclegan}.
While our method does not need any tuple supervision during the training stage and any input condition when generating new images, our model learns to generate images by sampling simple priors which can freely control each of disentangled factors. Hence, our model provides bidirectional diversity and generation ability along each of disentangled factors.

\vspace{0.1cm}
\noindent
\textbf{Generative adversarial networks}.
GANs have drawn wide attention in the community during the past few years \cite{GAN}.
GANs formulate a two-players min-max game where a discriminator learns to distinguish real samples from fake ones,
which are generated by a generator that tries its best to fool the discriminator. Many works extend the GANs framework to various generative applications, e.g., image generation \cite{DCGAN,pixcnn,COGAN}, image editing \cite{c59}, image-to-image translation \cite{pix2pix,cycleGAN,starGAN,UNIT,discogan} and variational inference \cite{aae}, etc.
Our proposed DSRGAN also extend the GANs framework to uniquely disentangle the underlying spatial structure
and the rendering without any tuple supervision.

Our framework is
related to CoupledGAN (CoGAN) \cite{COGAN} which also learns a pair of generative networks \cite{UNIT}. CoGAN learns to generate pairs of images in two domains with the same underlying structure, but it fails to generate images by separately controlling underlying structure and rendering.  The proposed DSRGAN are different in that we aim to explicitly disentangle underlying structure and rendering in a target domain. Hence, our model can control underlying structure and rendering of generated samples respectively.
\section{Methodology}

In this section we formulate our problem and illustrate the core idea of our method.

We aim to learn a target generator $G_t(z_s, z_{rt})$ that produces samples following $P_{t}(x_t)$, given images $\{x_t^i\}_{i=1}^{N_t}$  from the real data distribution $P_{t}(x_t)$ of target domain $\mathcal{D}_t$. Here underlying-structure prior $z_s\sim P_{z_s}(z_s)$  and rendering prior $z_{r}\sim P_{z_{rt}}(z_{rt})$  are expected to control underlying structure and rendering over underling structure, respectively.

Since this formulation is too unconstrained to disentangle the underlying spatial structure and rendering, we propose to introduce an auxiliary domain $\mathcal{D}_a$ to provide explicit guidance to the learning task. We assume that $\mathcal{D}_a$ to some extent shares a common underlying-structure space with $\mathcal{D}_t$ and has its specific rendering space.
In some cases, we can even invent a new auxiliary domain  $\mathcal{D}_a$ by manipulating $\mathcal{D}_t$ with simple low-cost image processing techniques, which is shown in our experiments.
\begin{figure}
\small
\centering
\includegraphics[width=0.75\linewidth]{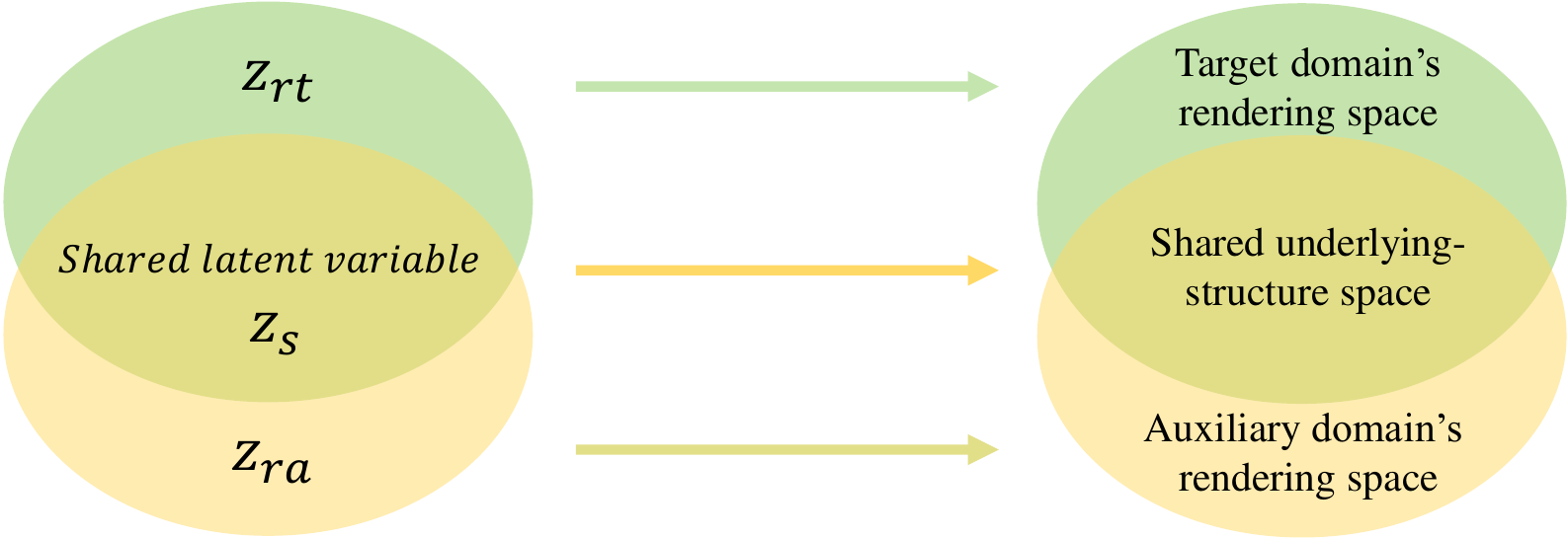}
\caption{
Illustration of critical idea.
In our model, the shared latent variable $z_s$ is expected to represent shared underlying spatial structures in both domains, and specific latent variables $z_{rt}$, $z_{ra}$ are expected to represent specific renderings, respectively. .  (Best viewed in color.)}
\label{fig:idea_2}
\vspace{-0.45cm}
\end{figure}

\begin{figure*}[t]
\small
\centering
\includegraphics[width=0.7\linewidth]{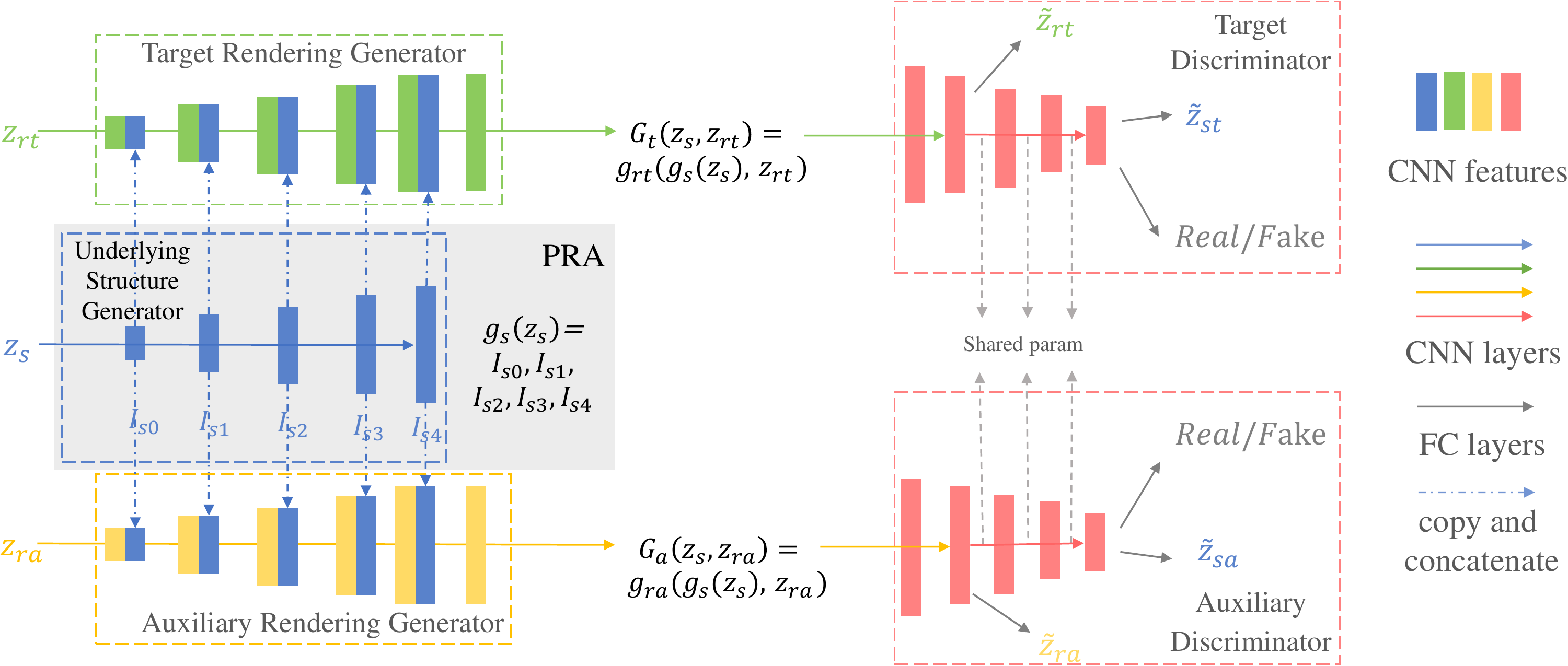}
\vspace{-0.05cm}
\caption{\label{fig:framework}
Illustration of the DSRGAN framework. Please compare this figure with Figure \ref{fig:idea_2}. Target generator $G_t$ and auxiliary generator $G_a$  contain rendering generator $g_{rt}$  and $g_{ra}$ respectively and share a common Progressive Rendering Architecture (PRA), which consists of two elements: (1) a shared underlying structure generator $g_s$ ; (2) the way to progressively provide underlying-structure information (CNN features generated by $g_s$) to $g_{rt}$  or $g_{ra}$. (Best viewed in color.)}
\vspace{-0.5cm}
\end{figure*}

We illustrate our main thought in Figure \ref{fig:idea_2}. In the figure,  domain $\mathcal{D}_t$ and $\mathcal{D}_a$  contain shared underlying-structure space. The latent variables of domain $\mathcal{D}_t$ and $\mathcal{D}_a$ are partially shared. The shared latent variable $z_s$ is expected to represent shared underlying spatial structures in both domains, and specific latent variables $z_{rt}$, $z_{ra}$ are expected to represent specific renderings of the two domains, respectively.  For the disentangled task, the key is to enforce a common latent variable $z_s$  only to represent shared underlying structure. It is equivalent to disentangling underlying structure and rendering. This is because images' factors can be just divided into underlying structure and rendering on underlying structure. When the partially shared input prior $z_s$ is enforced to represent underlying structure, the rest of input priors (i.e., domain-specific priors $z_{rt}$ and $z_{ra}$) will be unavoidably arranged to represent renderings, i.e., the presentations of  the two factors  can be successfully disentangled into $z_s$, $z_{rt}$ or $z_s$, $z_{ra}$ .

Therefore, in order to enforce $z_s$ to represent shared underlying structure, we introduce another parallel task which learns an auxiliary generator  $G_a(z_s, z_{ra})$ given $\{x_a^i\}_{i=1}^{N_a}$ from $P_{a}(x_a)$ of auxiliary domain $\mathcal{D}_a$. For that,  we propose to learn the parallel generators $G_t$ and $G_a$   by training a pair of GANs.
This joint adversarial learning task can be formulated as:
\vspace{-0.3cm}
\begin{small}
\begin{align}\label{eqn:loss_adv}
&\min_{G_t, G_a}\max_{D_t, D_a}\mathcal{L}_{adv} \\ \nonumber
  = & \mathbb{E}_{x_t\sim P_{t}}[\log(D_t(x_t))] + \mathbb{E}_{x_a\sim P_{a}}[\log(D_a(x_a))]  \\\nonumber
+ & \mathbb{E}_{z_s\sim P_{z_s}, z_{rt}\sim P_{z_{rt}}}[\log(1-D_t(G_t(z_s, z_{rt})))]\\\nonumber
+ & \mathbb{E}_{z_s\sim P_{z_s}, z_{ra}\sim P_{z_{ra}}}[\log(1-D_a(G_a(z_s, z_{ra})))],
\end{align}
\end{small}
where two GANs consist target generator $G_t$, target discriminator $D_t$ and auxiliary generator $G_a$, auxiliary discriminator $D_a$  respectively.
Specially,  we propose a novel framework DSRGAN which consists a pair of GANs to learn two parallel subtasks:  each GAN learns  to generate images in respective domains. In the training stage,  since two generators possess a common network (Underlying Structure Generator $g_s$), they tend to let the common network $g_s$ learn the common factor, i.e., shared underlying structure.  Further, we propose a Progressive Rendering Architecture based on $g_s$ which ulteriorly models the inherent relation of underlying structure and rendering. Next, we  elaborate pairs  of generators and discriminators in our proposed framework as illustrated in  Figure \ref{fig:framework}.


\vspace{0.1cm}
\noindent
\textbf{Generator}.
As shown in Figure \ref{fig:framework}, the target generator $G_t$  and the auxiliary generator $G_a$  possess a target rendering generator $g_{rt}$  and an auxiliary rendering generator $g_{ra}$  respectively and share a common \emph{\textbf{P}rogressive \textbf{R}endering \textbf{A}rchitecture} (\textbf{PRA}), which consists of two elements: (1) a shared underlying structure generator $g_s$ ; (2) the way to progressively provide underlying-structure information (CNN features generated by $g_s$) to rendering generator $g_{rt}$ or $g_{ra}$.
$z_s$,  $z_{rt}$  and $z_{ra}$  follow simple uniform distribution.

In our model, the generative process can be separated into two stages. The first stage is that PRA's underlying structure generator $g_s$ models common underlying structure and provides underlying-structure information to both rendering generators $g_{rt}$ and $g_{ra}$.  The second stage is that $g_{rt}$ and $g_{ra}$  generate rendering over underlying structure by processing underlying-structure information from $g_s$. It is similar to people's painting, i.e., drawing an skeleton firstly and then colorizing the skeleton.

We first elaborate PRA's first element, i.e., underlying structure generator $g_s$. As illustrated in Figure \ref{fig:framework}, both generators $G1$ and $G_a$  share the same $g_s$ with input of shared prior $z_s$. If the two generators have no weight-sharing constraint and are two totally independent network,  $G_t$ and $G_a$ would learn two independent marginal distributions , then $z_s$ would randomly represent domain $\mathcal{D}_t $'s and $\mathcal{D}_a$'s factors, which mostly have no relationship between each other. because $z_s$ is mapped to two image domains by two independent functions. However, our proposed $G_t$ and $G_a$  share a common network $g_s$. Hence, $g_s$ constrains behaviors of  both generators $G_t$  and $G_a$ . $G_t$ and $G_a$ are responsible to learn to generate images in two domain respectively, thus during training, both generators need to generate underlying structure.  As the shared network $g_s$ of two generators, $G_t$  and $G_a$ tend to enforce $g_s$ to learn shared factor of both domains, i.e., underlying structure. Thus, as input of $g_s$, shared prior $z_s$ is enforced to represent underlying structure. Further, target-domain special prior  $z_{rt}$ and auxiliary-domain specific prior $z_{ra}$  are unavoidably arranged to represent special renderings of two domains respectively, as image domains can be divided into two factors (underlying structure and rendering). Due to underlying generator $g_s$  which is able to model underlying structure, CNN features generated by $g_s$ can be treated as underlying-structure information,  which is provided to two rendering generators $g_{rt}$ and $g_{ra}$.

Next, we elaborate target rendering generator $g_{rt}$  and PRA's second element, i.e., the way to progressively provide information to $g_{rt}$ . Since our framework is symmetric, the counterpart $g_{ra}$ can be similarly defined. For natural images, rendering should be generated over underlying spatial structure. For example, when generating an image of blue sneakers with red shoelaces, the spatial distribution of the two colors is very fine-grained, and heavily depends on the exact shape of the sneakers. Thus, rendering should be generated by the guidance of spatial information.
Hence, we propose the PRA's way to progressively provide underlying-structure information to $g_{rt}$. As illustrated in Figure \ref{fig:framework}, the intermediate CNN features $I_{s0}$ to $I_{s4}$ generated by $g_s$ are progressively inputted to $g_{rt}$. This progressive way is to guide every layer of $g_{rt}$ to generate more precise intermediate rendering information for aligning underlying structure, so that $g_{rt}$ can finally generate fine-grained rendering to better match underlying structure. However, if we only input features of one layer from $g_s$, the rendering may match the very fine-grained spatial distribution badly, as several layers of $g_{rt}$ may generate bad intermediate information of rendering without the guidance of the additional spatial information. Meanwhile, as another input of $g_{rt}$, latent prior $z_{rt}$ is encouraged to control the rendering generation.
 

As analyzed before, the Progressive Rendering Architecture (PRA) can enforce shared latent prior $z_s$ to represent underlying structure due to underlying structure generator $g_s$ of it and it also validly models the inherent relationship between underlying structure and rendering due to the way to progressively provide information.

\vspace{0.1cm}
\noindent
\textbf{Discriminator}. As illustrated in the right of Figure \ref{fig:framework},
two discriminators $D_t$ and $D_a$  give the probability that a given sample is from target domain or auxiliary domain. Their last several layers are weight-sharing. Besides considering to reduce the model's parameters, this design also take into consideration that the first several layers process low-level image information \cite{2014_NIPS_transferable} which is
mostly corresponding to the specific rendering, while the latter shared layers mainly process the high-level image information \cite{COGAN}, i.e., shared underlying spatial structure.

Let us consider another problem in our method, i.e., a trivial solution is to ignore $g_s$ 's generation ability and thus $g_{rt}$ and $g_{ra}$  take charge of the whole learning tasks and so that the input $z_s$  controls little information in generated images, and vice versa.  To prevent prior noises $z_s$ ,$z_{rt}$  and $z_{ra}$  from being ignored, we add a loss
\cite{infoGAN} for reconstructing all noises:
\vspace{-0.15cm}
\begin{small}
\begin{align}
\label{eqn:lip}
\small
&\mathcal{L}_{ns} = \mathbb{E}_{z_s\sim P_s, z_{rt}\sim P_{rt}, z_{ra}\sim P_{ra}}[ \\ \nonumber
\small
&\mu_1(||z_s - \tilde{z}_{st}|| + ||z_s - \tilde{z}_{sa}||) + \mu_2(||z_{rt} - \tilde{z}_{rt}|| + ||z_{ra} - \tilde{z}_{ra}||)],
\end{align}
\end{small}
where $\mu_1$, $\mu_2$ control relative importance.
As shown in Figure \ref{fig:framework}, $z_{rt}$ and $z_{ra}$  are reconstructed from the second specific layer of  two discriminators by one full-connected layer respectively, and $z_s$ is reconstructed from the last layer by one full-connected layer, as rendering and underlying structure are low-level and high-level information respectively.
We also add another loss to regularize our model. As two discriminators can predict  $\tilde{z}_{st}$ $ and $ $\tilde{z}_{rt}$ or $\tilde{z}_{sa}$ and $\tilde{z}_{ra}$ from real images $x_t$ or $x_a$, so that two generators can reconstruct the real images $x_t$ or $x_a$ with $\tilde{z}_{st}$ and $\tilde{z}_{rt}$ or $\tilde{z}_{sa}$ and $\tilde{z}_{ra}$ . This is because in optimality, generated samples follow the same distribution as real ones \cite{GAN}. $\mathcal{L}_{rec}$ is a reconstruction loss as:
\vspace{-0.15cm}
\begin{small}
\begin{align}
\label{eqn:lrec}
\mathcal{L}_{rec} = \mathbb{E}_{x_t\sim P_{t}, x_a\sim P_{a}}[ ||x_t - \tilde{x}_{t}|| + ||x_a - \tilde{x}_{a}|| ],
\end{align}
\end{small}
where $\tilde{x}_{t}$ and $\tilde{x}_{a}$ are images reconstructed from  real images.

\vspace{0.1cm}
\noindent
\textbf{Full Objective}.
Our full loss is formulated as:
\vspace{-0.15cm}
\begin{small}
\begin{align}
\label{eqn:loss}
\mathcal{L} =&\mathcal{L}_{adv}
+\lambda_1 \mathcal{L}_{ns} + \lambda_2 \mathcal{L}_{rec},
\end{align}
\end{small}
where $\lambda_1$, $\lambda_2$ control relative importance.
\section{Quantitative Criteria for Disentanglability}\label{sec:evaluations}
We empirically evaluate DSRGAN by experiments on different datasets in Section \ref{sec:experiments}. In this section , we
elaborate quantitative criteria for \emph{disentanglability} and provide intuitive understanding for our criteria.
We define the disentanglability as the ability to disentangle the underlying spatial structure and the rendering on the structure in a model for disentangling, so that $z_s$ and $z_r$ can \emph{fully and only} control the variation of each of two disentangled factors, respectively.

However, a proper quantitative measure of disentanglability is missing in current literature as far as we know.
To quantify the concept of ``fully and only'', we define two difference functions $d_s(x_a, x_b)$ and $d_r(x_a, x_b)$ where $x_a$, $x_b$ are images,
$d_s: \mathcal{X}\times\mathcal{X}\rightarrow [0, 1]$ measures the variation/difference of underlying spatial structures between $x_a$ and $x_b$,
while $d_r: \mathcal{X}\times\mathcal{X}\rightarrow [0, 1]$ measures the difference of renderings.
Both the diversity along each of the two dimensions (underlying structure and rendering) and the independence across the two dimensions should be taken into consideration. Thus, we can quantify the disentanglability  by Normalized Disentanglability (ND) :
\vspace{-0.2cm}
\begin{small}
\begin{align}
\label{eqn:disentanglability}
& ND = \mathbb{E}[\Delta d_s] + \mathbb{E}[\Delta d_r] \\ \nonumber
& = \mathbb{E}_{z_s, z_s'\sim P_s; z_r, z_r'\sim P_r}[ \\ \nonumber
&(d_s(G(z_s, z_r), G(z_s', z_r)) - d_s(G(z_s, z_r), G(z_s, z_r'))) \\ \nonumber
           + & (d_r(G(z_s, z_r), G(z_s, z_r')) - d_r(G(z_s, z_r), G(z_s', z_r)))],
\end{align}
\end{small}
where given different $z_s$, $z_s'$ and a fixed $z_r$, a successful model should have
high $d_s(G(z_s, z_r), G(z_s', z_r))$ (i.e., $z_s$  should control the underlying spatial structure as \emph{fully} as possible so that the variation of underlying structure should be as big as possible with varying $z_
s$) and low $d_s(G(z_s, z_r), G(z_s', z_r))$ (i.e., $z_r$ should control the underlying spatial structure as little as possible so that the variation of underlying structure should be as small as possible with varying $z_
r$),
and it is similar to $d_r(G(z_s, z_r), G(z_s, z_r'))$ and $d_r(G(z_s, z_r), G(z_s', z_r))$. Hence, $\mathbb{E}[\Delta d_s]$ measures a model's ability to use $z_s$  but not $z_r$ fully and only to control underlying structure and  $\mathbb{E}[\Delta d_r]$ measures model's ability to use $z_r$  rather than $z_s$ to fully and only control rendering. For example, if  a model has no ability to disentangle underlying structure, i.e., $z_s$ and $z_r$ have similar ability to control underlying structure, we have $(d_s(G(z_s, z_r), G(z_s', z_r)) = d_s(G(z_s, z_r), G(z_s, z_r')))$, i.e. $\mathbb{E}[\Delta d_s] = 0$, and similarly for $\mathbb{E}[\Delta d_r]$.Thus,  ND, i.e., $ \mathbb{E}[\Delta d_s] + \mathbb{E}[\Delta d_r]$, can reflect ability to disentangle.

\begin{figure}[t!]\centering\small
\subfigure[{ND=0.47 $\mathbb{E}[\Delta d_r]$=0.29 $\mathbb{E}[\Delta d_s]$=0.18
}]{
\includegraphics[width=0.2\linewidth]{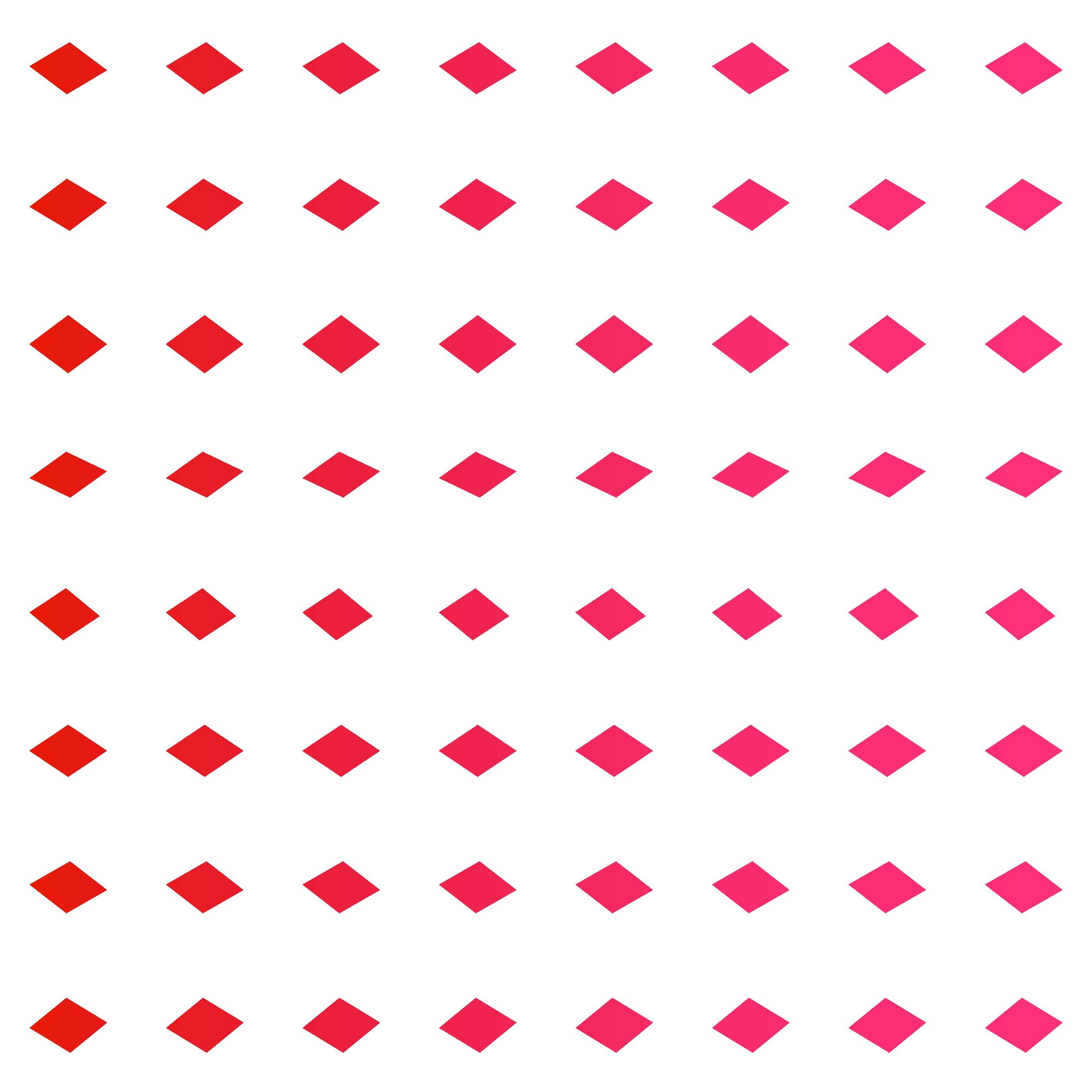}
}
\subfigure[{ND=0.88 $\mathbb{E}[\Delta d_r]$=0.71 $\mathbb{E}[\Delta d_s]$=0.18
}]{
\includegraphics[width=0.2\linewidth]{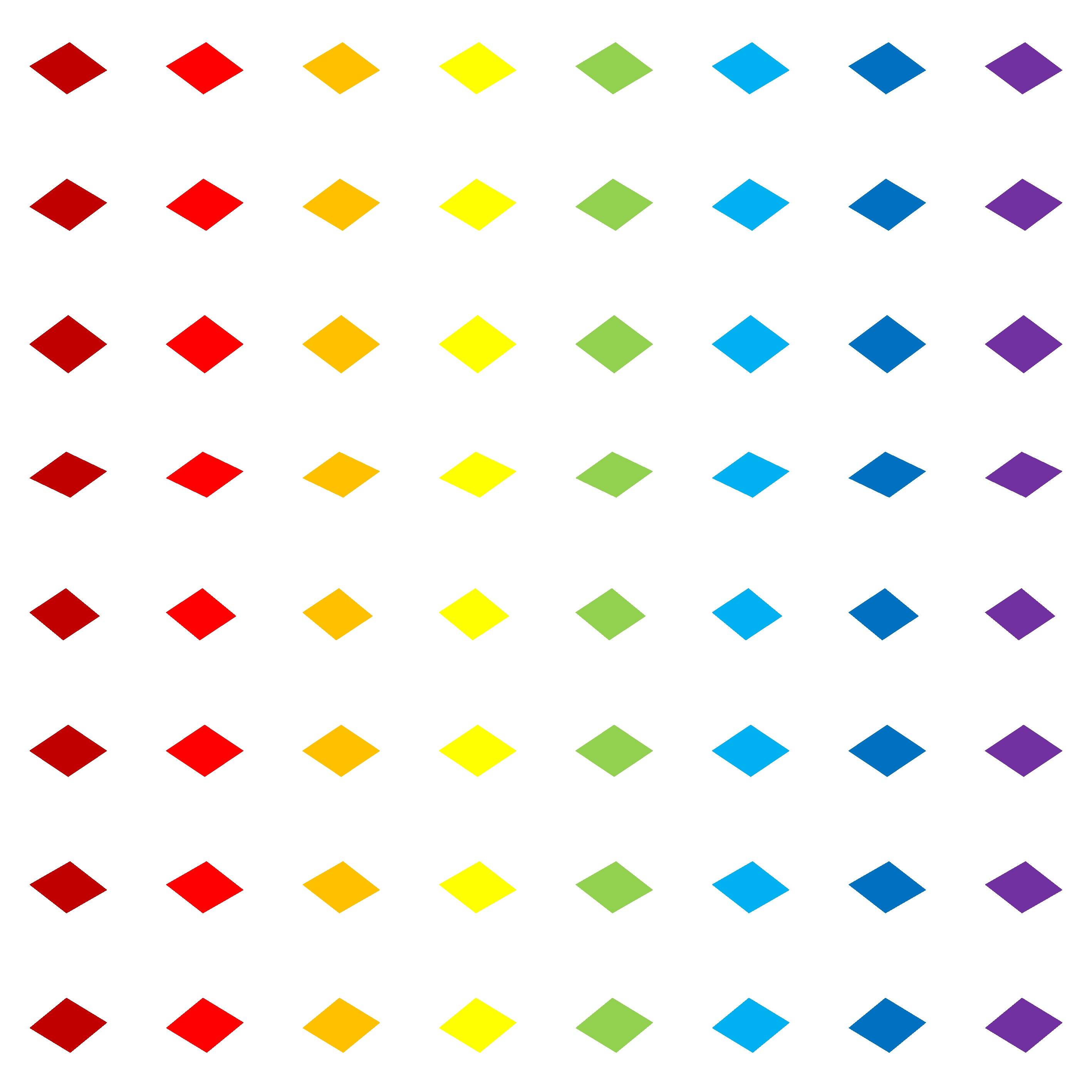}
}
\subfigure[{ND=0.82 $\mathbb{E}[\Delta d_r]$=0.29 $\mathbb{E}[\Delta d_s]$=0.53
}]{
\includegraphics[width=0.2\linewidth]{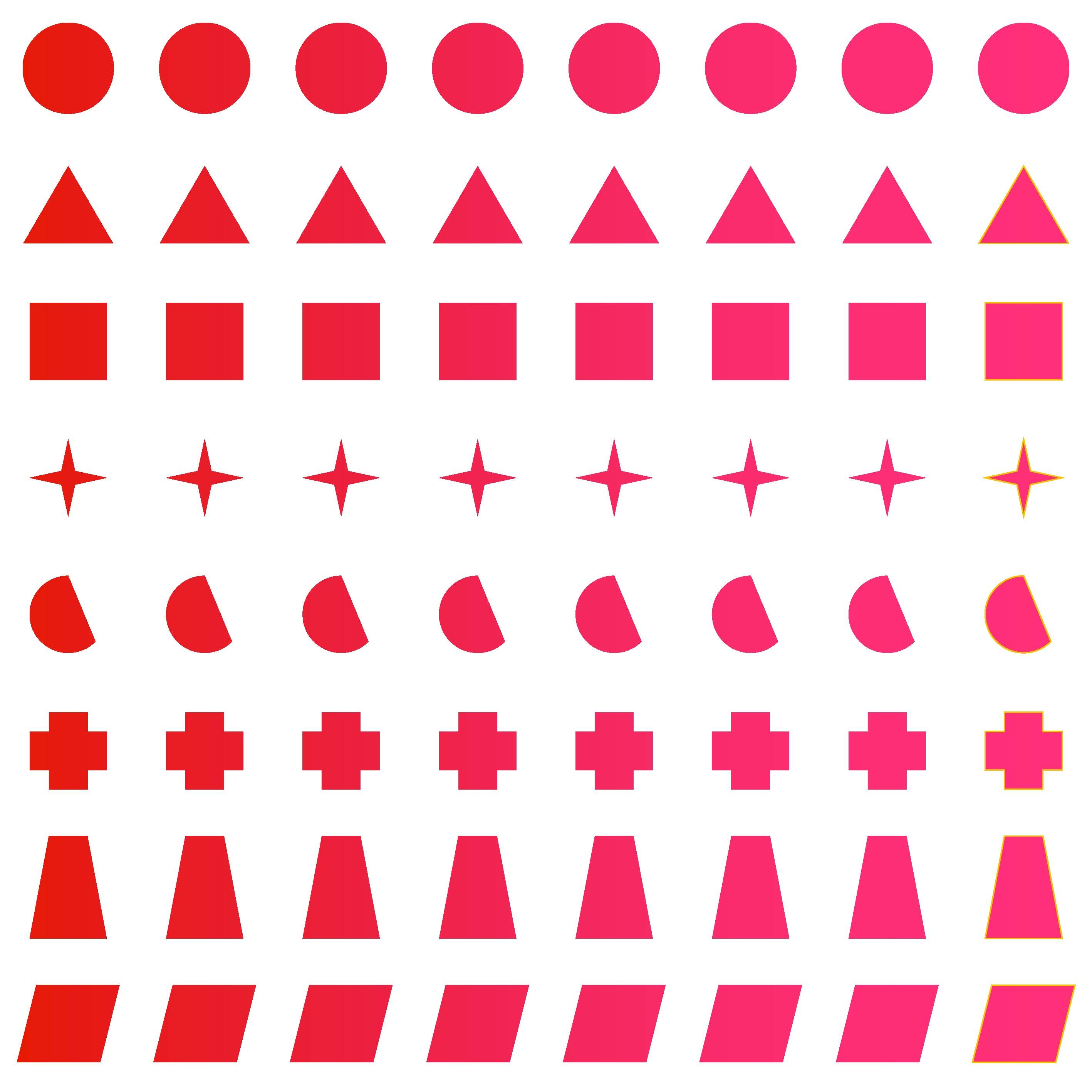}
}
\subfigure[{ND=1.24 $\mathbb{E}[\Delta d_r]$=0.71 $\mathbb{E}[\Delta d_s]$=0.53
}]{
\includegraphics[width=0.2\linewidth]{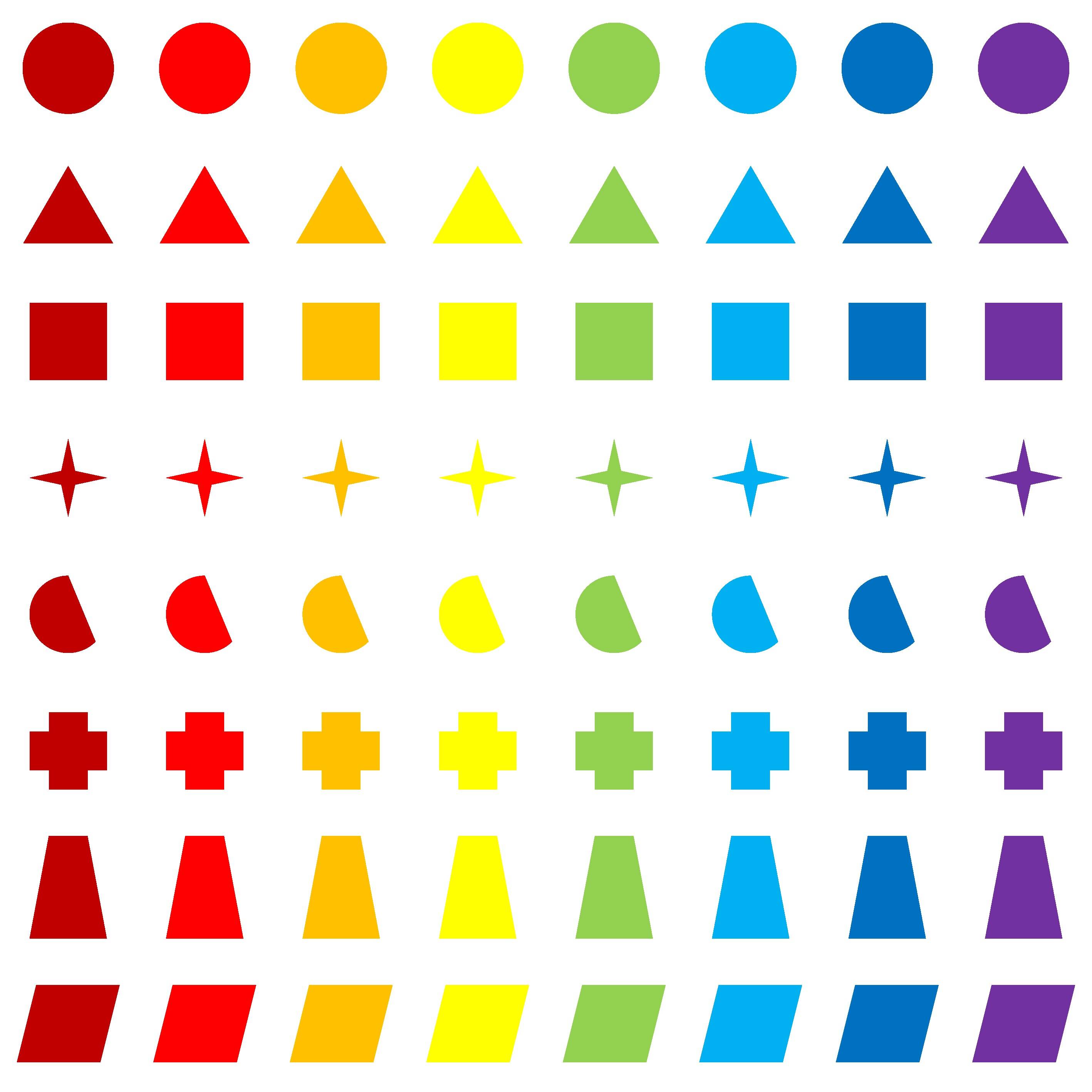}
}
\vspace{-0.1cm}
\caption{ 4 synthetic toy figures to show intuitive understanding for ND, $\mathbb{E}[\Delta d_s]$ and $\mathbb{E}[\Delta d_r]$. Please zoom in for better visualization.}
\label{fig:toy}
\vspace{-0.45cm}
\end{figure}
As shown in Figure \ref{fig:toy}, we synthesize 4 toy pictures to present intuitive understanding.
We assume that 4 subfigures are generated by 4 models. Images in every row are generated by fixing $z_s$ and varying $z_r$ randomly and images in every column are generated by fixing $z_r$ and  varying $z_s$ randomly. We use ND, $\mathbb{E}[\Delta d_s]$  and $\mathbb{E}[\Delta d_r]$ to evaluate on 4 subfigures ( where $d_r$ is the histogram distance to measure diversity of color, and $d_s$ is the FBPD distance to the diversity of shape. The two distances are elaborated in Section 5.1's quantitative results of disentanglability).  Comparing (a) and (b), we can see that $\mathbb{E}[\Delta d_r]$ is higher in (b), indicating that (b) has richer color diversity. Comparing (a) and (c), it can be seen that $\mathbb{E}[\Delta d_s]$ also reflects the richer diversity in (c) according to its higher value. Finally, since (d) has both advantages of (b) and (c), its ND is the highest among all subfigures. Hence,  ND, $\mathbb{E}[\Delta d_s]$ and $\mathbb{E}[\Delta d_r]$ agree with observation very well (we also show intuitive understanding for ND, $\mathbb{E}[\Delta d_s]$ and $\mathbb{E}[\Delta d_r]$ by real samples generated by real models in Appendix E).

\begin{table*}
\centering
\small
\caption{
\label{tbl:disentanglability}
Quantitative results. ``conditional-InfoGAN''  refers to providing the binary domain label to InfoGAN's noise input and discriminator input as condition. Explanation of the metrics can be found in Eq. (\ref{eqn:disentanglability}).
In the last row, ``real'' means the quantities are computed over the real data. $\mathbb{E}[d_s]$ is the averaged $d_s$ of all pairs of real data samples,
so that it can be regarded as the upper bound of $\mathbb{E}[\Delta d_s]$.
Similarly, $\mathbb{E}[d_r]$ can be regarded as the upper bound of $\mathbb{E}[\Delta d_r]$.}
\begin{tabular}{c|c|c|c|c|c|c|c|c|c}
\hline
\multirow{2}*{Method} & \multicolumn{3}{c|}{Task on shoes images } & \multicolumn{3}{c|}{Task on shoes images} & \multicolumn{3}{c}{Task on face images} \\
 & \multicolumn{3}{c|}{($d_s$ using HOG)} & \multicolumn{3}{c|}{($d_s$ using FBPD)} & \multicolumn{3}{c}{} \\
\cline{2-10}
& ND & $\mathbb{E}[\Delta d_r]$ & $\mathbb{E}[\Delta d_s]$ & ND & $\mathbb{E}[\Delta d_r]$ & $\mathbb{E}[\Delta d_s]$ & ND & $\mathbb{E}[\Delta d_r]$ & $\mathbb{E}[\Delta d_s]$ \\
\hline
\hline
InfoGAN                 &0.36&0.05&0.31&0.40  &  0.05  & 0.35 & 0.39 & 0.07 & 0.32  \\
$\beta$-VAE             &0.56& 0.13&0.43& 0.48 & 0.13   & 0.36 & 0.39 & 0.13 & 0.25  \\
\hline
conditional-InfoGAN     &0.41& 0.17&0.24& 0.46  & 0.17  & 0.29  & 0.43 & 0.11 & 0.32   \\
conditional-$\beta$-VAE &0.59& 0.10&\textbf{0.49}& 0.46 & 0.10  & 0.36 & 0.41 & 0.14 & 0.27   \\
SD-GAN                  &0.49& 0.15&0.34& 0.55  & 0.15  & 0.40 & / & / &/ \\
\hline
Ours                    &\textbf{0.65}&\textbf{0.24}&0.41&\textbf{0.66}  & \textbf{0.24}  &  \textbf{0.42}  &\textbf{0.65} & \textbf{0.29}& \textbf{0.36}  \\
\hline
Real & /&$\mathbb{E}[d_r]$=0.66&$\mathbb{E}[d_s]$=$0.58 $&/ & $\mathbb{E}[d_r]$=0.66  & $\mathbb{E}[d_s]$=0.43 & / & $\mathbb{E}[d_r]$=0.64 & $\mathbb{E}[d_s]$=$0.51$ \\
\hline
\end{tabular}
\vspace{-0.6cm}
\end{table*}
\section{Experiments}\label{sec:experiments}
In this section, we will present our three tasks, in two of which, for different tasks, we design specific $d_s$ and $d_r$ . The implementation details of DSRGAN are given in Appendix D. Next, we first introduce the datasets, tasks and compared alternative methods, then we analyze the experimental results in Section 5.1  and 5.2. Some further evaluations on our model is presented by ablation study in Section 5.3. 

\vspace{0.1cm}
\noindent
\textbf{Datasets and Tasks}.
In the first group of experiments we use the shoes images dataset available from \cite{pix2pix} which contains $50,025$ images of various shoes.
In this dataset we have one task, where the underlying spatial structure and rendering refer to the shape and color style of the shoes, respectively.
We use the original dataset as $\mathcal{D}_t$, and manually invent a $\mathcal{D}_a$ by simply transforming all the samples into grayscale images.
In this way, the specific renderings in Figure \ref{fig:idea_2} now refer to the color style and the ``grayscale style'', respectively, and the shared underlying spatial structure refers to the shape.
We show some samples of all tasks (including every $\mathcal{D}_t$ and $\mathcal{D}_a$) we used in Appendix A.
In the second group of experiments we use the CelebA dataset \cite{celeba}, which contains $202,599$ celebrity face images. We have two tasks in CelebA, including disentangling
(1) human identity and hair color
and (2) human identity and whether wearing a pair of glasses.

\vspace{0.1cm}
\noindent
\textbf{Alternative Methods}.
In our experiments we compare DSRGAN with other three alternative disentangled generative models,
including a supervised model SD-GAN \cite{SDGAN} which uses tuple supervision and two unsupervised models InfoGAN \cite{infoGAN} and $\beta$-VAE \cite{betaVAE}.
As InfoGAN and $\beta$-VAE do not explicitly distinguish the disentangled image's factors,
in each of their tasks, we plot along all latent dimensions to pick a best-disentangled dimension (against the remaining dimensions)
for comparison, following \cite{betaVAE}.
Note that as our model implicitly uses the binary ``domain label'', for a fair quantitative comparison we also provide the domain information to the compared unsupervised models, i.e. infoGAN and $\beta$-VAE, by providing the binary domain label for both noise input and discriminator input as ``conditions'' \cite{cGAN}.
We denote the resultant models as conditional-InfoGAN and conditional-$\beta$-VAE, respectively.

\subsection{Results on the Shoes Images Dataset}\label{sec:shoes_results}
As the rendering we concern in this task is the color style, for our model we only show the results generated by learning from target domain, i,e., colorful shoes.

\begin{figure}[t]\small
\centering
\subfigure[DSRGAN, $G_t$]{
\includegraphics[width=0.305\linewidth]{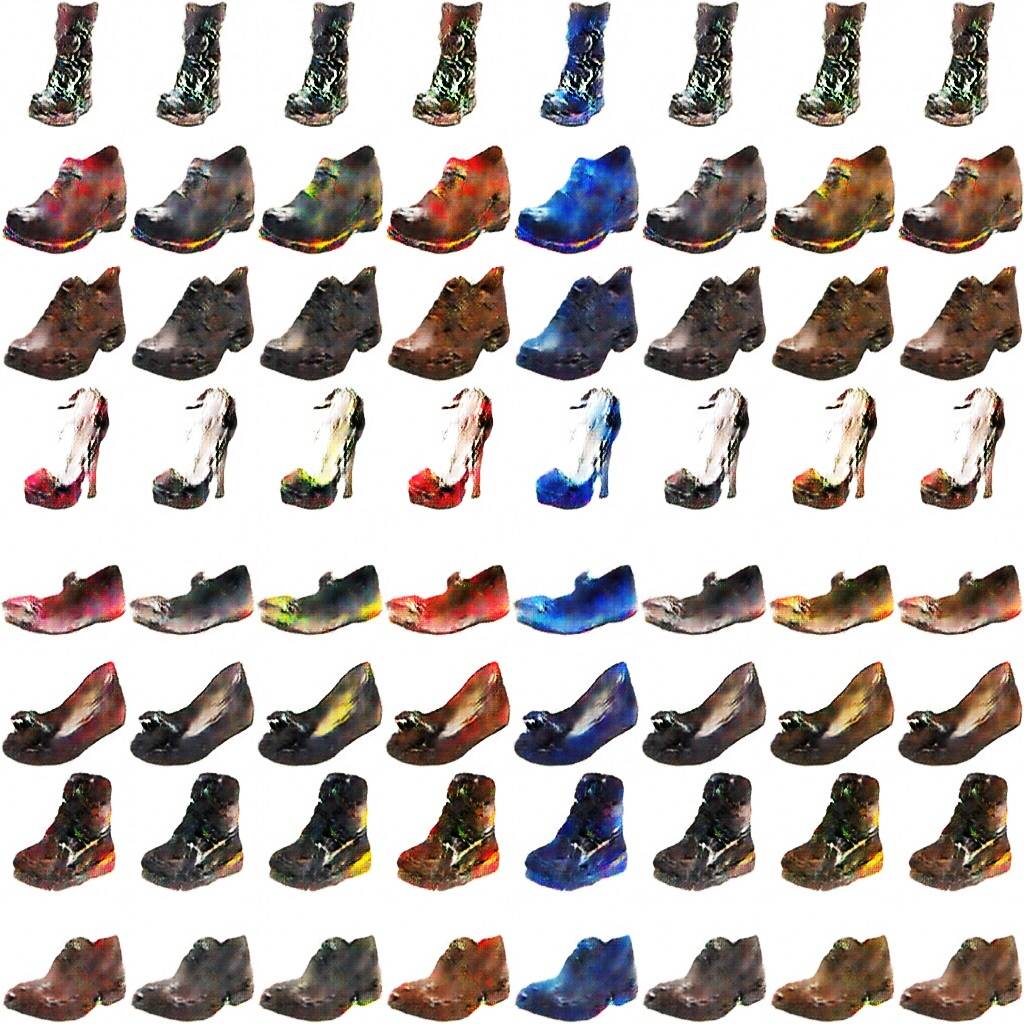}
}
\subfigure[SD-GAN]{
\includegraphics[width=0.305\linewidth]{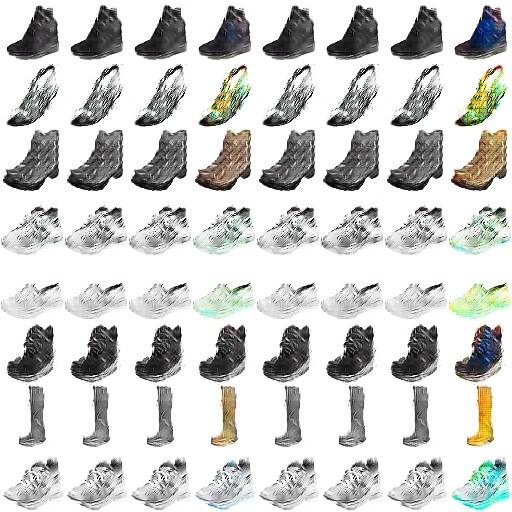}
}
\subfigure[InfoGAN]{
\includegraphics[width=0.305\linewidth]{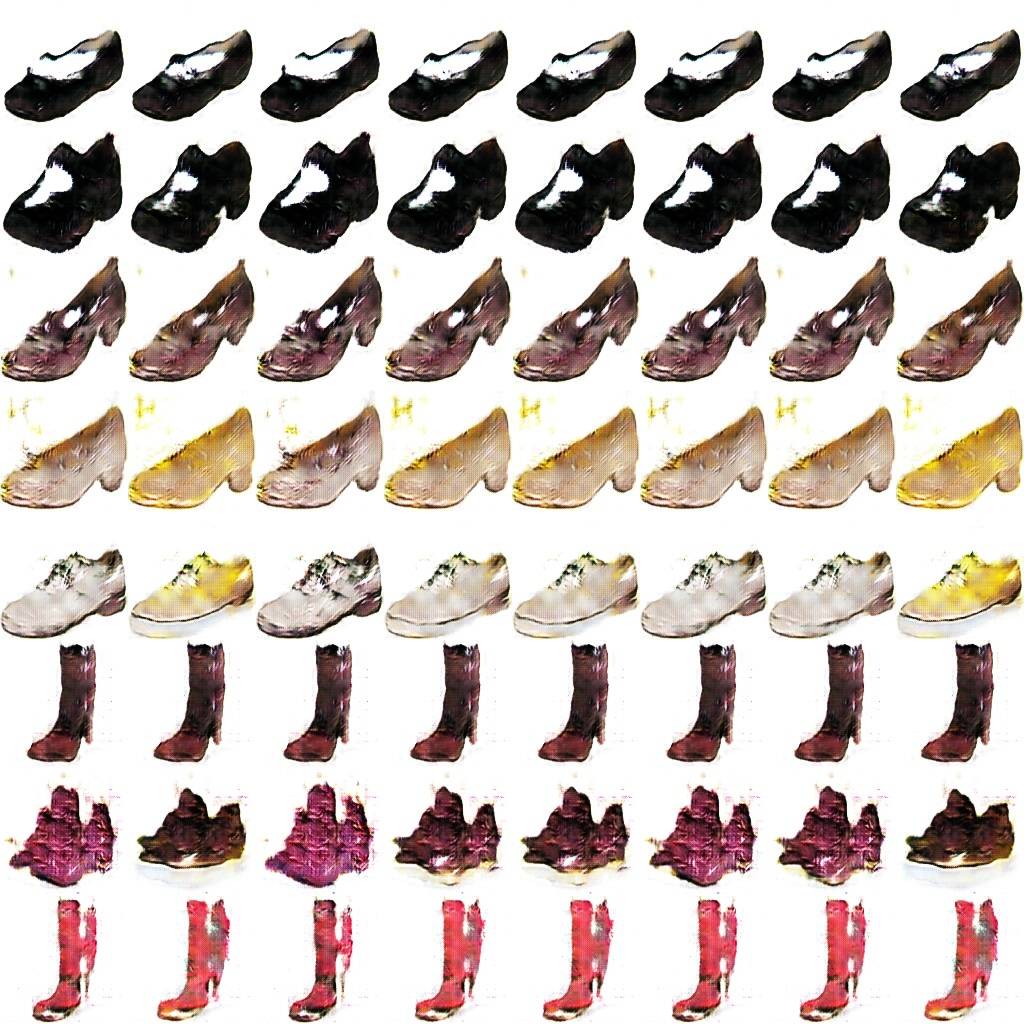}
}
\subfigure[$\beta$-VAE]{
\includegraphics[width=0.305\linewidth]{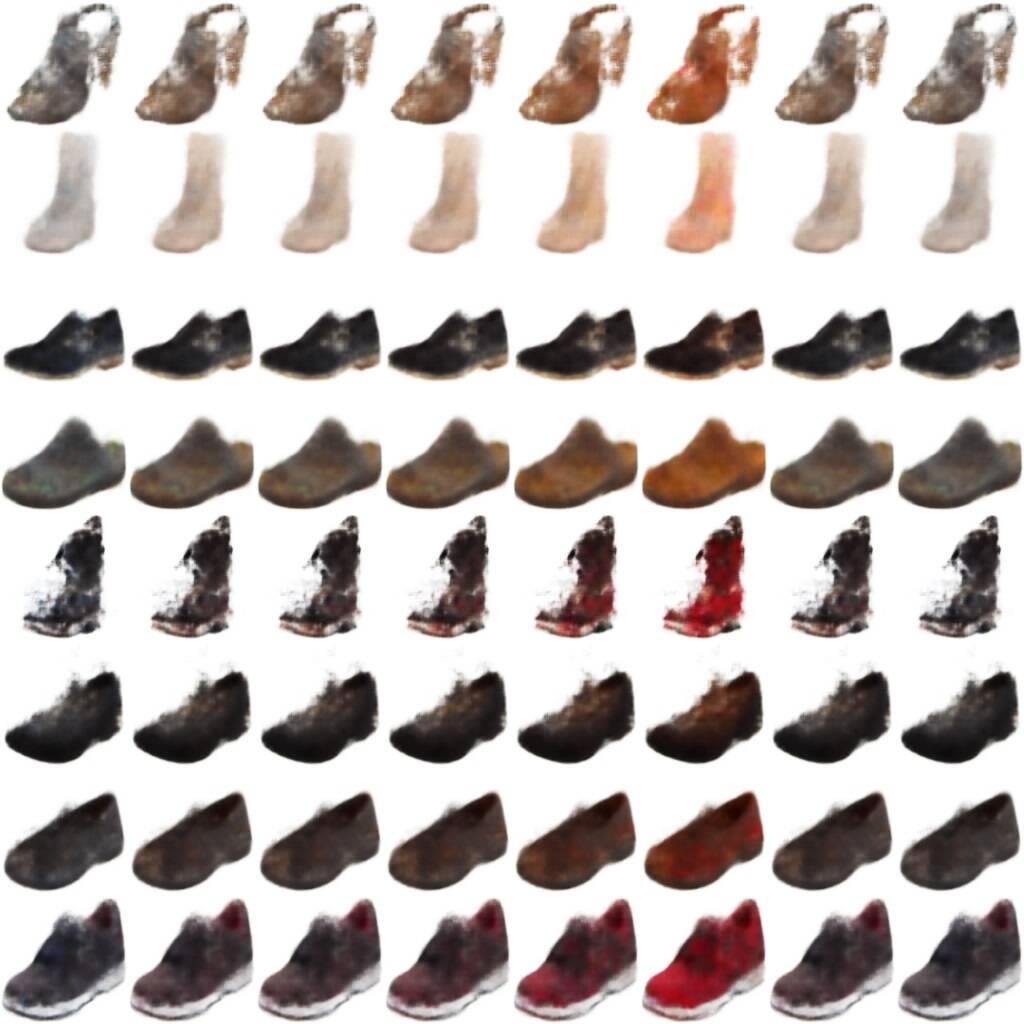}
}
\subfigure[\scriptsize{Conditional-InfoGAN}]{
\includegraphics[width=0.305\linewidth]{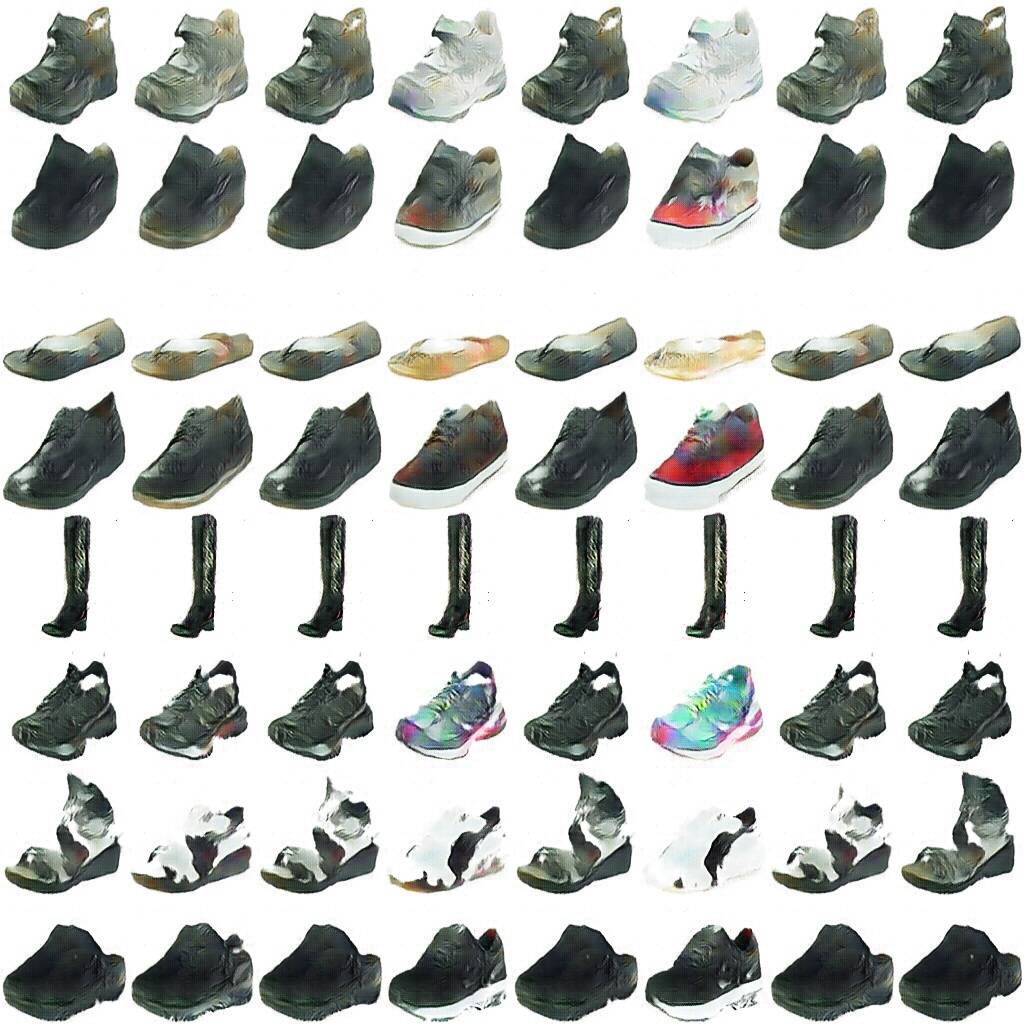}
}
\subfigure[Conditional-$\beta$-VAE]{
\includegraphics[width=0.305\linewidth]{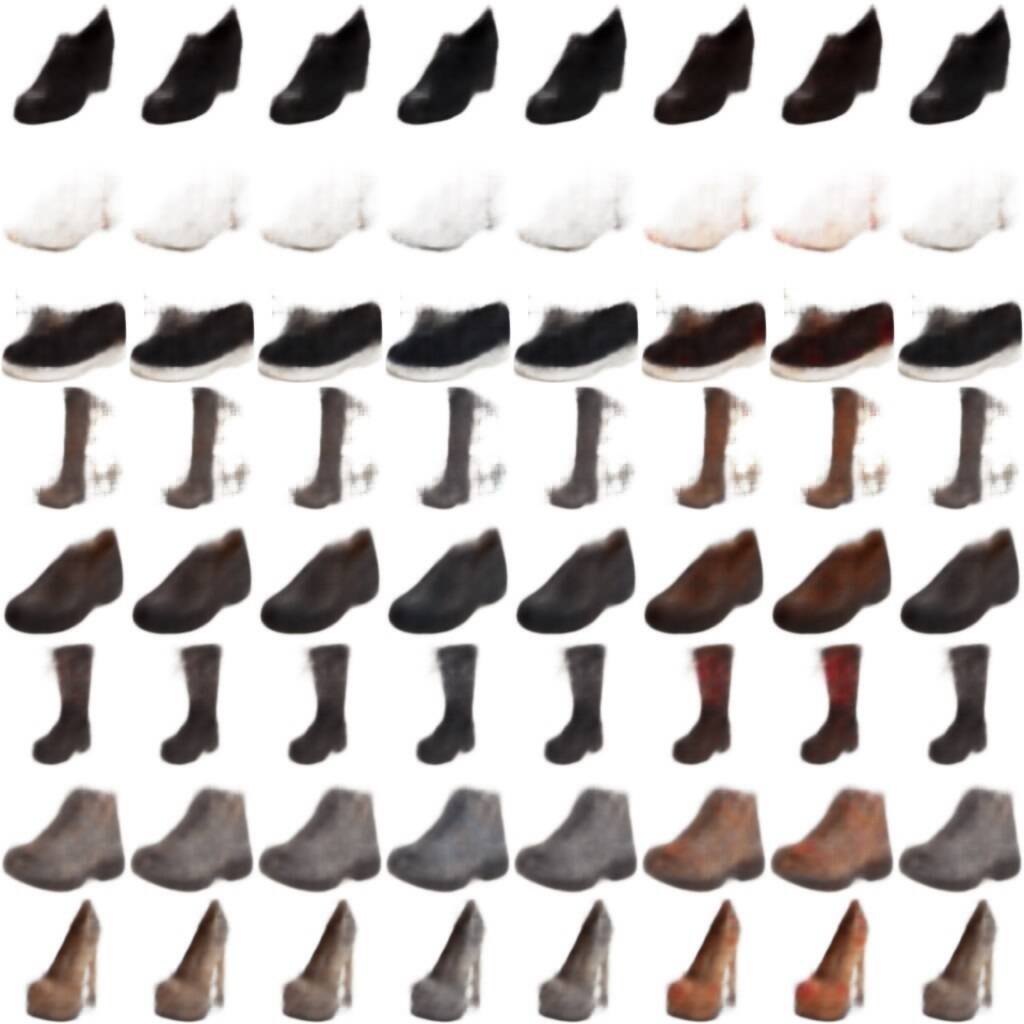}
}
\caption{Visualization results in disentangling the shapes and color styles of the generated shoes images. \textbf{Samples in each row are generated with the same underlying-structure noise $z_s$ but randomly sampled different rendering noise $z_r$. Samples in each column are generated with the same $z_r$ but randomly sampled different $z_s$. This presentation format is also used in Figure \ref{fig:visual_comparison_hair} and Figure \ref{fig:visual_comparison_glasses}.} Please zoom in for better visualization.}
\label{fig:visual_comparison_shoes}
\vspace{-0.45cm}
\end{figure}

\vspace{0.1cm}
\noindent
\textbf{Qualitative results}.
We first show some representative visual results in Figure \ref{fig:visual_comparison_shoes}.
\textbf{More visual results of all tasks can be found in Appendix B, including results on an additional dataset which contains 138,767 images of handbags \cite{pix2pix}.}
We can see that our DSRGAN successfully disentangles the shapes and the color styles of the generated images. Specifically, the diversity along each dimension and the independence across the two dimensions can be clearly observed. As shown in Figure \ref{fig:visual_comparison_shoes}, shoes in the same column are generated by sampling from  different $z_s$ and the same $z_r$, and similarly, shoes in the same row are generated by sampling from  the same $z_s$ and different $z_{r}$. Our generated shoes in same row look the same in aspect of underlying structure and possess colorful rendering. Similarly,  shoes in the same column possess similar color  and various outline. However, in other models' results,  shoes generated by sampling the same  $z_s$ and different $z_r$ ( shoes in the same row) possess different outlines, e..g., InfoGAN and conditional-infoGAN's shoes, or possess low diversity in rendering, e.g., shoes generated by $\beta$-VAE and conditional-$\beta$-VAE are less colorful than ours.
\vspace{0.1cm}
\noindent

\textbf{Quantitative results of disentanglability}.
In this task the underlying spatial structure and the rendering to be disentangled refer to shape and color style of the shoes, respectively.
Therefore, using different $z_s,z_s'$ should result in a great difference in generated shapes and a slight difference in color, i.e. high $d_s(x_a, x_b)$ and low $d_r(x_a,x_b)$.
To quantify the shape difference, we define $d_s(x_a, x_b)$ in Eqn. \ref{eqn:disentanglability} in two ways:

(1) Histogram of Oriented Gradients (HOG) \cite{HOG} for quantifying local shape. The distance of HOG is used to measure the difference of edges and curves within local windows between two images. We use the default setting in the original paper \cite{HOG}. After extracting HOG features for $x_a$ and $x_b$, we use the normalized Euclidean distance.

(2) Foreground Binary Pixel Disagreement (FBPD) for quantifying general contour shape.
Since in the shoe dataset the background is always white, we transform an image to grayscale and simply regard all pixels in the range of $[250, 255]$ as background. This empirically works well, confirmed by our visual inspection.
Then the disagreed points can be counted, and we can get
the ratio of the area of non-overlapping foregrounds in $x_a$ and $x_b$, over the area of their foregrounds union.

Then, we define $d_r$ which quantifies the color style difference as the normalized Euclidean distance of $hist(x_a)$ and $hist(x_b)$, where $hist(\cdot)$ is the color histogram \cite{hist} in HSV space which well models the human perception on color \cite{HSV}. We set the bin sizes to [18, 8, 8] corresponding to the three channels. Hue channel has 18 bins because the color varies mostly in this channel. Therefore, there are in total $18*8*8=1152$ fine-grained color ranges (if too many it will not be robust).
To estimate the expectation, we use $10,000$ pairs of $z_s$/$z_s'$ and $z_r$/$z_r'$.
We show the comparative results in the left part of Table \ref{tbl:disentanglability}.

From Table \ref{tbl:disentanglability} we can see that our model achieves the highest disentanglability in both quantitative measures.
The main reasons are as follows:
Compared to SD-GAN which uses pairwise supervision but lacks a component to associate the two GANs,
DSRGAN contains a specifically designed model architecture to capture the shared latents for disentangling.
While Conditional-infoGAN, infoGAN, Conditional-$\beta$-VAE and $\beta$-VAE maximize the mutual information between images and the priors, they lack a principled mechanism to explicitly disentangle specific factors, and thus their results are somewhat random.
In contrast, we use an auxiliary domain and design a novel architecture based on which we can explicitly disentangle the shared underlying structure and the specific renderings.
\vspace{0.1cm}
\noindent

\begin{figure}[t]\small
\centering
\subfigure[DSRGAN, $G_t$]{
\includegraphics[width=0.305\linewidth]{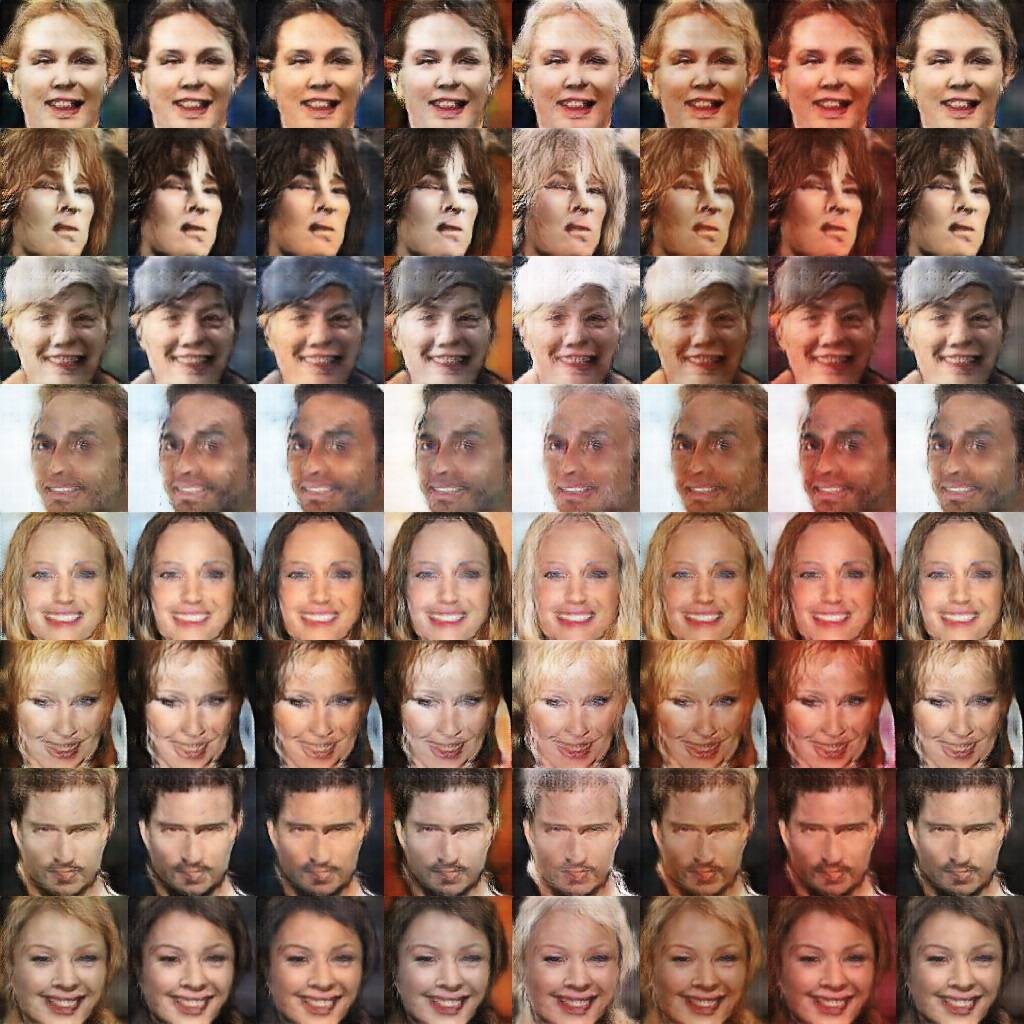}
}
\subfigure[DSRGAN, $G_a$]{
\includegraphics[width=0.305\linewidth]{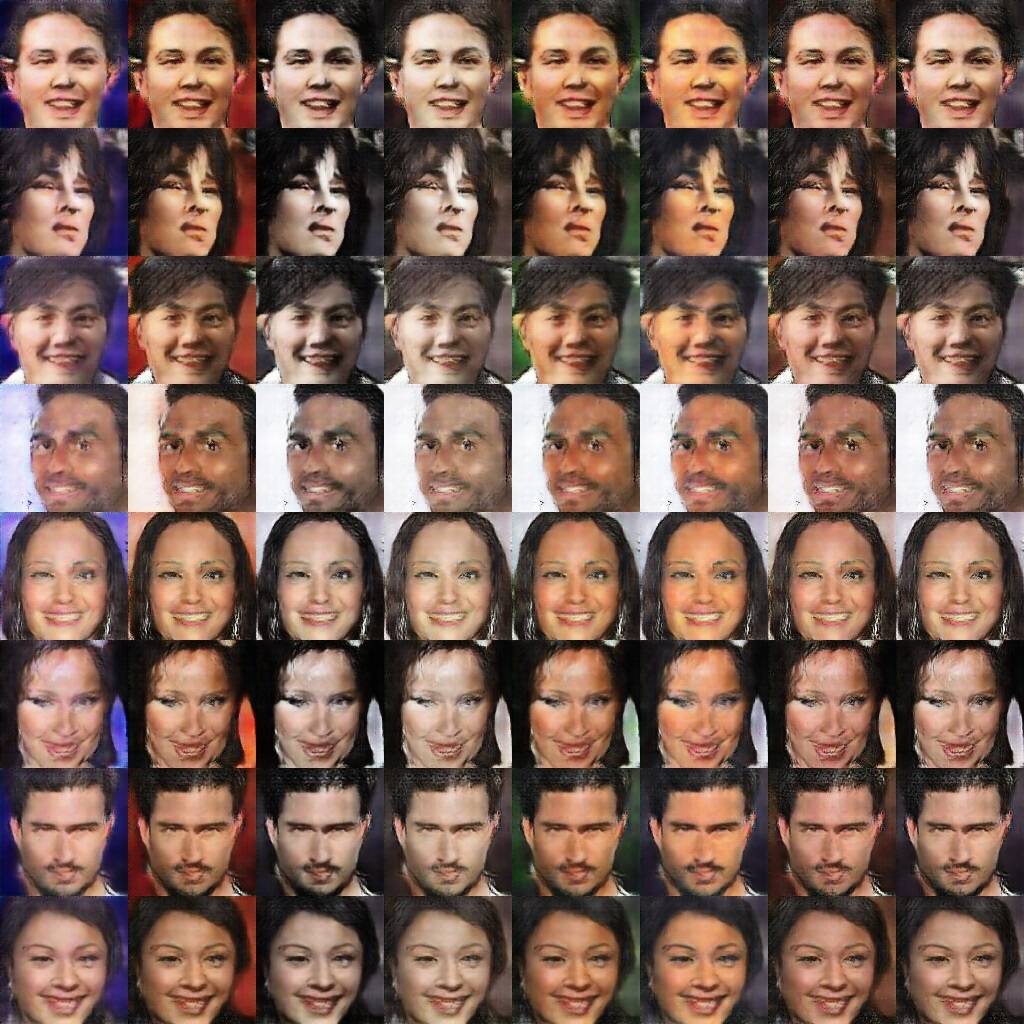}
}
\subfigure[InfoGAN]{
\includegraphics[width=0.305\linewidth]{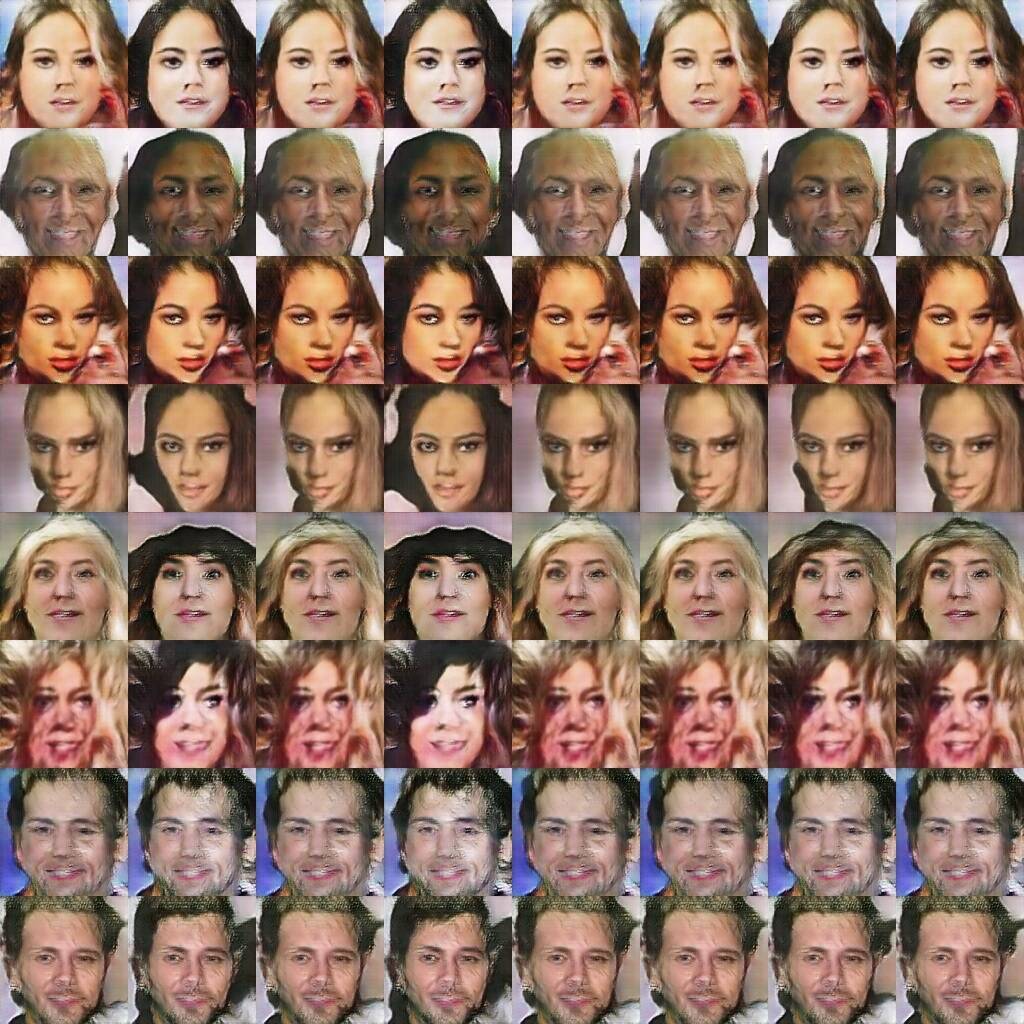}
}
\subfigure[$\beta$-VAE]{
\includegraphics[width=0.305\linewidth]{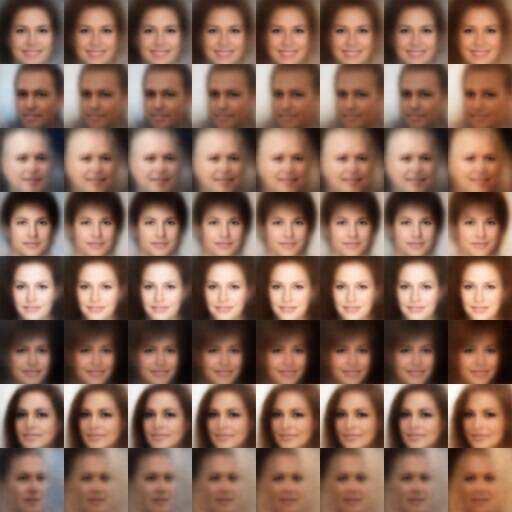}
}
\subfigure[\scriptsize{Conditionl-InfoGAN}]{
\includegraphics[width=0.305\linewidth]{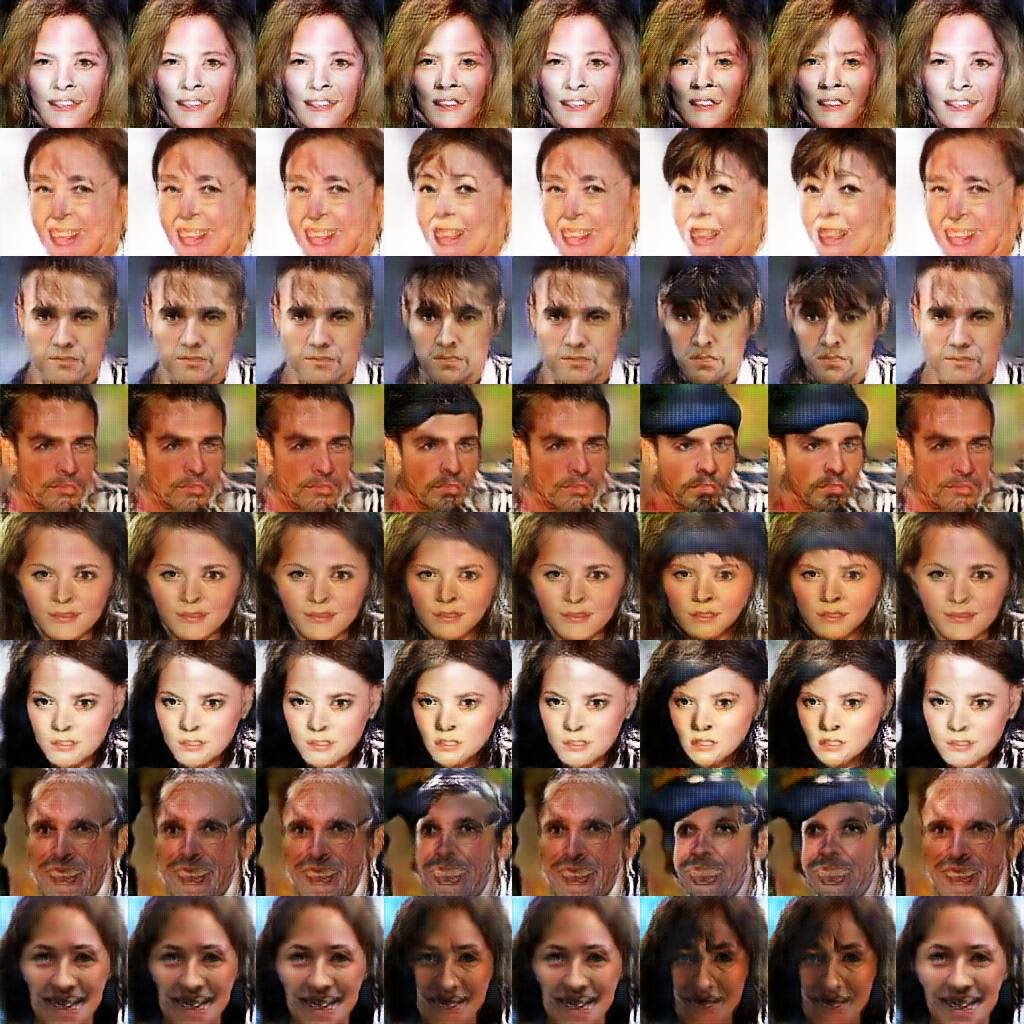}
}
\subfigure[Conditional-$\beta$-VAE]{
\includegraphics[width=0.305\linewidth]{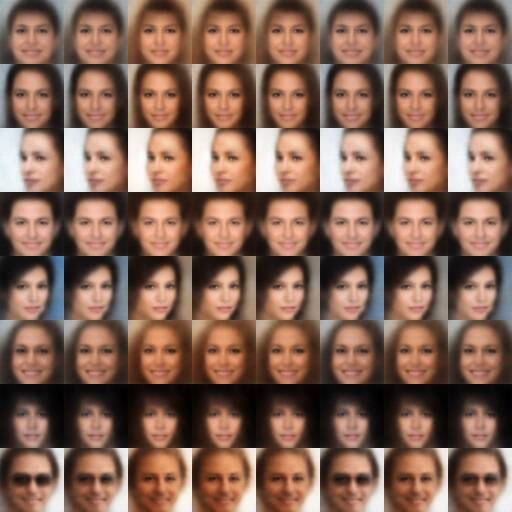}
}
\caption{Results in  disentangling the human identity and hair color of the generated face images.}
\label{fig:visual_comparison_hair}
\vspace{-0.2cm}
\end{figure}
\begin{table}[t!]\small
\centering
\caption{Success Rate (SR) in the shoes task}
\begin{tabular}{c|c|c|c|c}
\hline
 & InfoGAN & $\beta$-VAE & SD-GAN & Ours \\
\hline
SR &$91.2\%$ &$95.3\%$ &$89.4\%$ & $\textbf{95.6\%}$ \\
\hline
\end{tabular}
\label{tbl:qua_shoes}
\vspace{-0.45cm}
\end{table}

\textbf{Quantitative results of image quality}.
For image quality, we organized $50$ human annotators to judge whether a generated sample successfully resembles a real one.
We show some typical successful cases and failure cases in Appendix C, which were provided to annotators as reference.
For each method, an annotator judges $500$ samples and thus we can compute a success rate of $50*500=25,000$ samples.
We show the results in Table \ref{tbl:qua_shoes}.
We can see that our method is comparable to or better than the alternative methods in terms of quality of generated images.
This is in accordance with our visual inspection.
We note that such comparability in quality is also observed in the following tasks.
\textbf{As we do not claim superior quality,
in the following, we only focus on disentanglability}.

\vspace{0.1cm}
\noindent

\subsection{Results on the Face Images Dataset}\label{sec:face_results}

\textbf{Disentangling human identity and hair color}.
In the first task in this dataset, we aim to disentangle human identity (underlying structure) and hair color (rendering),
given:

\vspace{0.1cm}
\noindent
Target domain $\mathcal{D}_t$: images of people each of whom can have specific hair color (e.g. John has blond hair and Jane has red hair);

\noindent
Auxiliary domain $\mathcal{D}_a$: images of people whose hair color is black, i.e. no diversity in hair color.

\vspace{0.1cm}
\noindent
 We divide CelebA dataset into $D_t$ and $D_a$ according to the hair color labels. Both domains share a common underlying-structure space, since the subjects of the images in both domains are all human faces (although different identities).
Examples can be found in Appendix A. Note that in CelebA no pair-wise label is available,
so we could not train SD-GAN.
We show the visual results in Figure \ref{fig:visual_comparison_hair}.
We can see that $G_t$ successfully learns to disentangle the human identity and hair color, as faces generated by the same $z_s$ (images in the  same row )  possess the same ID and multiple hair colors and  faces generated by the same $z_r$ (images in the same column )  possess multiple IDs and similar hair colors. $G_a$ cannot generate faces with colorful hairs, since, in $\mathcal{D}_a$, the diversity in hair color is missing, while it can generate faces with the  same ID. Furthermore, faces generated by sampling the same $z_s$ in the same row between Figure \ref{fig:visual_comparison_hair} (a) and (b) share the same ID, i.e., shared underlying structure of both domains.

We also perform \textbf{quantitative evaluation on disentanglability}.
We define $d_s(x_a, x_b)$ as the normalized Euclidean distance of $f(x_a)$ and $f(x_b)$ where $f(\cdot)$ is a deep feature extractor, i.e. FaceNet, trained in a face recognition dataset \cite{facenet}, so that the features are identity-discriminative.
And we define $d_r$ as the normalized Euclidean distance of $hist^{1/3}(x_a)$ and $hist^{1/3}(x_b)$,
where $hist^{1/3}(\cdot)$ is the color histogram on the upper one third of the image,
because here we roughly regard the upper part as the region of hair.
As the results shown in the right part in Table \ref{tbl:disentanglability},
our model also significantly outperforms alternative methods in this task.

\vspace{0.1cm}
\noindent
\textbf{Disentangling human identity and whether wearing glasses}.
In this task the rendering refers to whether wearing a pair of glasses,
given:

\vspace{0.1cm}
\noindent
Target domain $\mathcal{D}_t$: images of people some of whom wear glasses while the others do not;
\noindent
Auxiliary domain $\mathcal{D}_a$: images of people that do not wear glasses.

\begin{figure}[t]\small
\centering
\subfigure[DSRGAN, $G_t$]{
\includegraphics[width=0.305\linewidth]{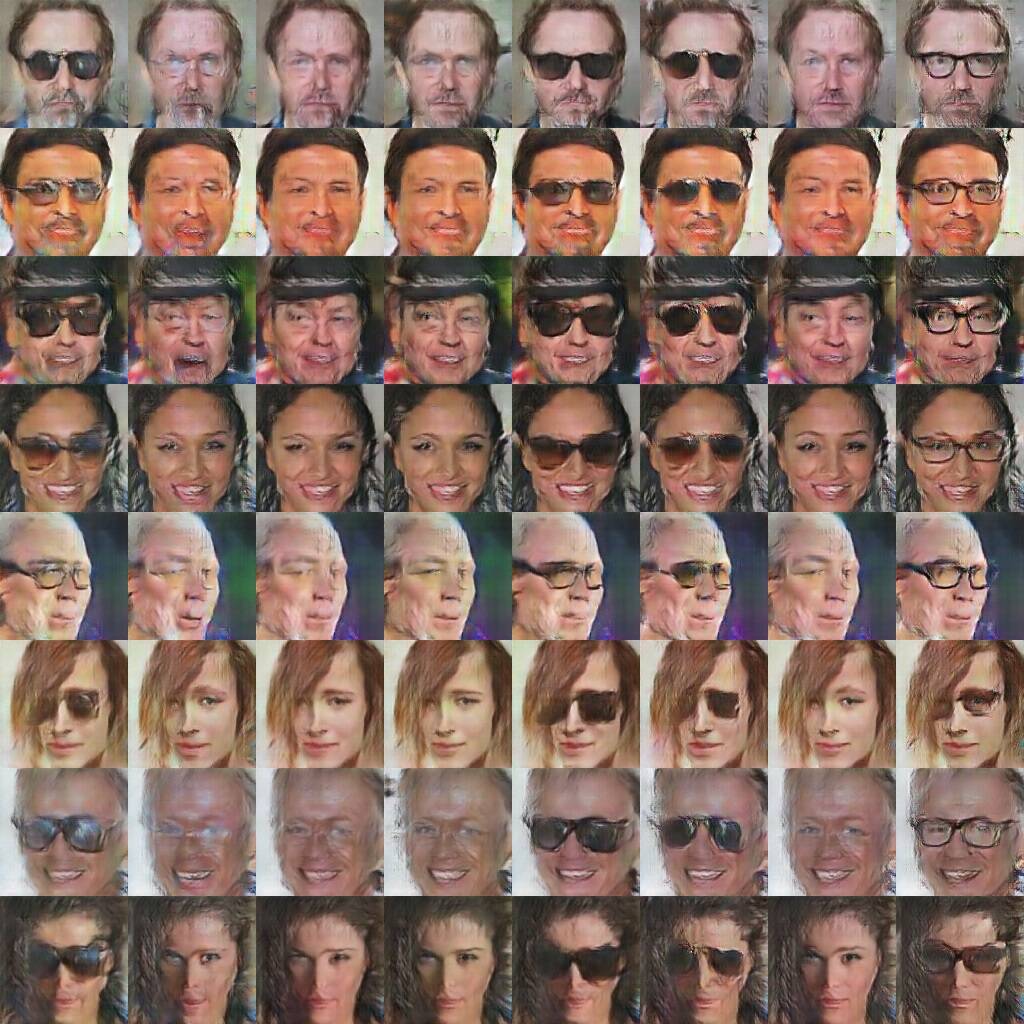}
}
\subfigure[DSRGAN, $G_a$]{
\includegraphics[width=0.305\linewidth]{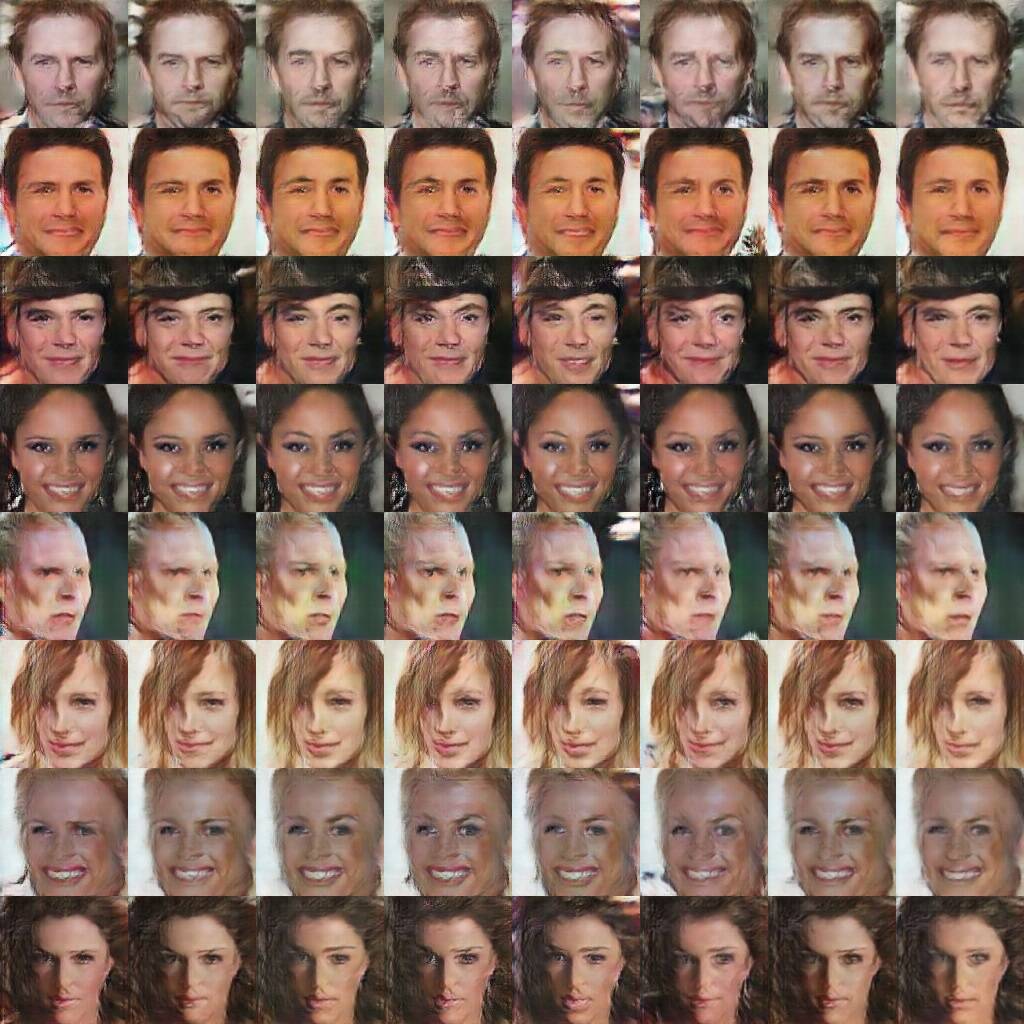}
}
\subfigure[InfoGAN]{
\includegraphics[width=0.305\linewidth]{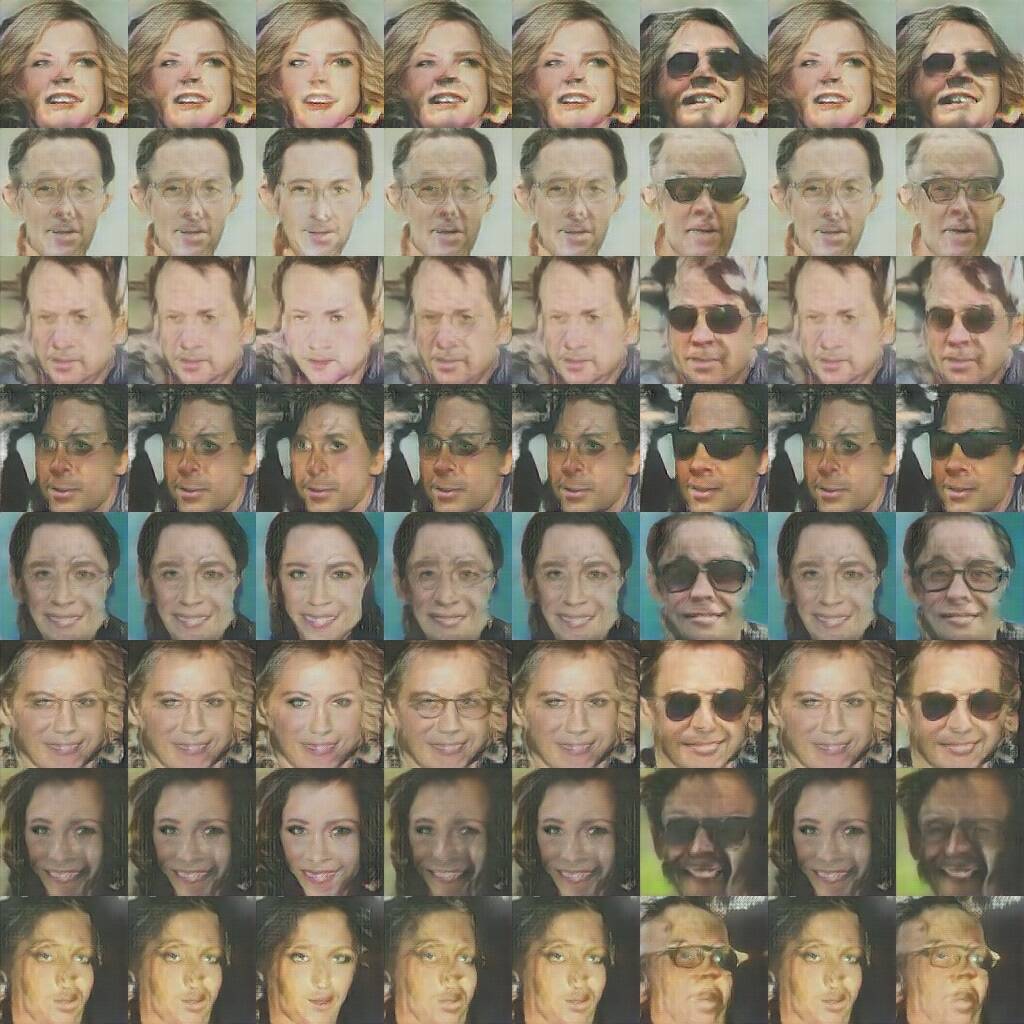}
}
\subfigure[$\beta$-VAE]{
\includegraphics[width=0.305\linewidth]{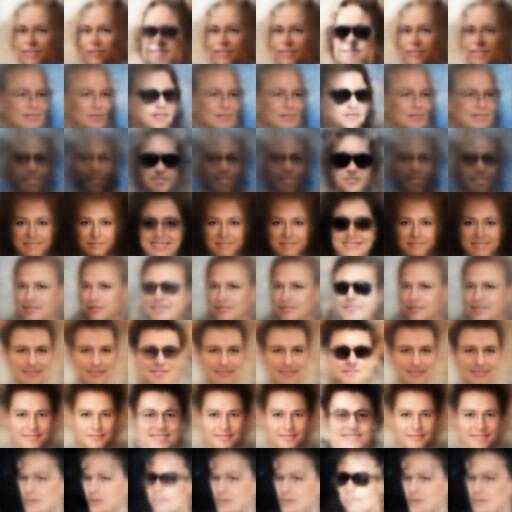}
}
\subfigure[\scriptsize{Conditional-InfoGAN}]{
\includegraphics[width=0.305\linewidth]{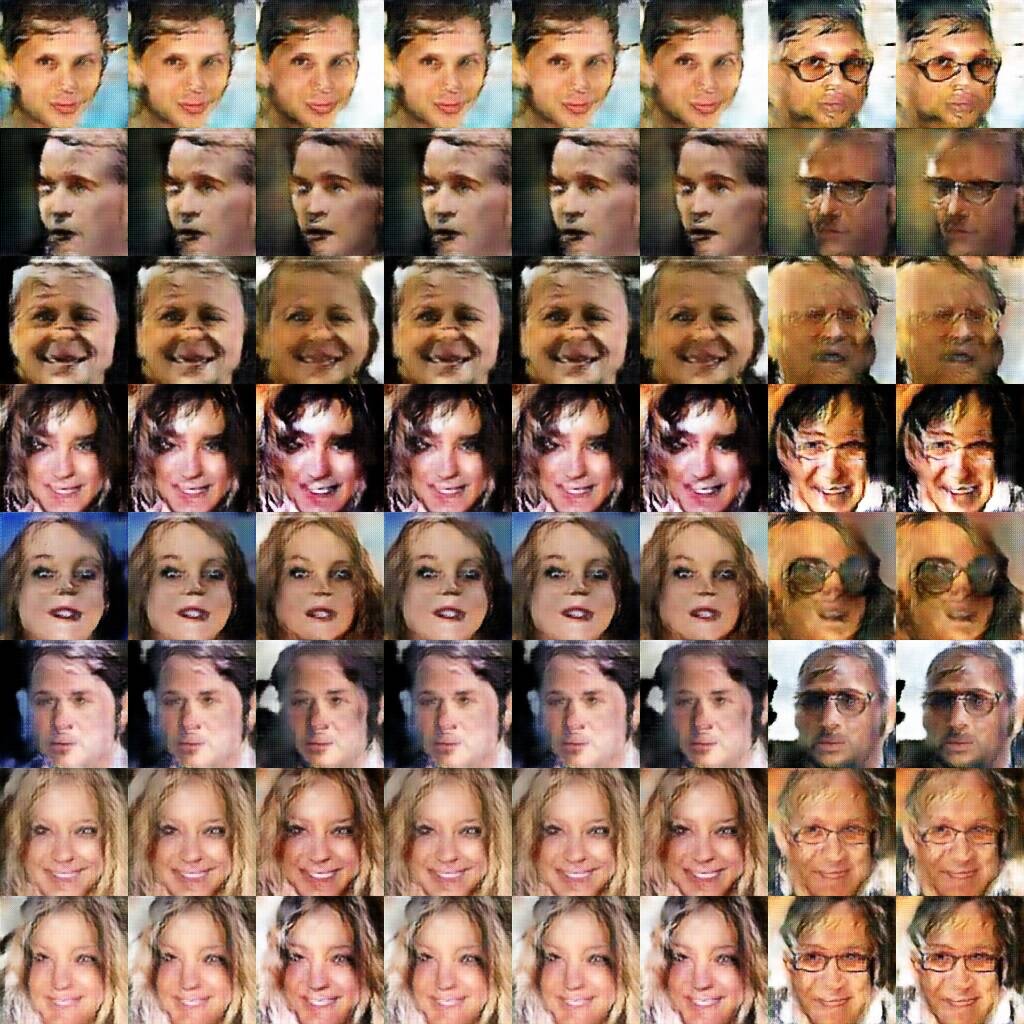}
}
\subfigure[Conditional-$\beta$-VAE]{
\includegraphics[width=0.305\linewidth]{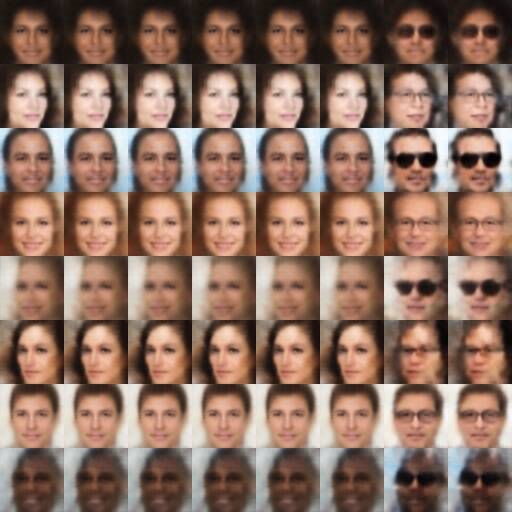}
}
\caption{Results in disentangling the human identity and wearing glasses of the generated images.}
\vspace{-0.2cm}
\label{fig:visual_comparison_glasses}
\end{figure}

\vspace{0.1cm}
\noindent
Similar to the last task, we split the dataset and perform dimension selection for compared methods.
We show our visual results in Figure \ref{fig:visual_comparison_glasses} (a). Compared to other models, our generated faces by sampling the same $z_s$ (i.e., faces in the same row) can possess more similar appearance and more various eyeglasses, e.g.,
sunglasses and transparent eyeglasses.

\begin{table}[t]\small
\caption{
\label{tbl:ablation}
Full ablation study in the task of disentangling the human identity and hair color.
Please refer to the Ablation Study section for detailed explanation of model variants.
``full'' refers to the
full model. }
\scriptsize
\setlength{\tabcolsep}{0pt}
\begin{tabular}{c|c|c|c|c|c|c|c|c}
\hline
Ablation & \multicolumn{2}{c|}{Auxiliary} & \multicolumn{2}{c|}{Progressive Rendering}&Shared & \multicolumn{2}{c|}{Auxiliary}\\
on which & \multicolumn{2}{c|}{domain}& \multicolumn{2}{c|}{Architecture (PRA)}&parameters& \multicolumn{2}{c|}{losses} \\
\hline
Model & w/o  & w/o auxiliary& w/o the PRA's& w/o  & w/o shared& w/o &  w/o  & \\
variants& auxiliary &domain \&   & progressive& PRA & parameters & $\mathcal{L}_{ns}$ & $\mathcal{L}_{rec}$ & full \\
&  domain & retaining PRA & way &&of $D_t \& D_a$ &  & &  \\
\cline{2-4}
\hline
        ND &0.37&0.58&0.52&0.37&0.62&0.63&0.63&0.65\\
\hline
$\mathbb{E}[\Delta d_r]$ &0.33&0.20&0.10&0.35&0.23&0.23&0.32&0.29 \\
$\mathbb{E}[\Delta d_s]$  &0.04&0.38&0.42&0.02&0.39&0.41&0.31&0.36 \\

\end{tabular}
\vspace{-0.6cm}
\end{table}

\subsection{Ablation Study}\label{sec:further_evaluations}
We report the ablation study results in the task of disentangling the human identity and hair color in Table \ref{tbl:ablation}, where the ablation components are  auxiliary domain, Progressive Rendering Architecture (PRA), shared parameters of discriminators,  and auxiliary losses.

 \textbf{Effect of Auxiliary domain}. As illustrated in the top of Figure \ref{fig:framework},  considering PRA is designed for disentangling with the help of auxiliary domain, removing the auxiliary domain means just keeping target rendering generator and target discriminator and inputting $z_s$ and $z_{rt}$ to target rendering generator together. ND changes from full model's 0.65 to w/o auxiliary domain's 0.37, i.e.,  the ability to disentangle drops by  $43 \%$ compared to that of the full model. It indicates the auxiliary domain is very essential to our model, as auxiliary domain can provide explicit guidance to the task for disentangling underlying structure and rendering.
When the auxiliary domain is removed and the PRA is retained, the performance (i.e., ND of w/o auxiliary domain $\&$ retaining PRA)  increases by $57\%$ compared to that of w/o auxiliary domain. This is because PRA well models the inherent relationship between underlying structure and rendering so that the disentanglablity can be markedly increased with the effect of PRA when there is no auxiliary domain.

\textbf{Effect of Progressive Rendering Architecture (PRA)}. Considering PRA contain two key element: (1) underlying structure generator $g_s$; (2)the way to progressively provide features generated by $g_s$ to two rendering generators.  We firstly just remove the PRA's progressive way by removing all the concatenations except last -layer features from $g_s$ in Figure \ref{fig:framework} and there is still $g_s$, the performance drops by $20 \%$ compared to the full model's. This is because rendering needs to be generated over underlying structure and our proposed architecture could well model the spatial dependence of renderings on the underlying spatial structures. When we remove whole PRA further (i.e., removing $g_s$) and input $z_s$ with $z_{rt}$ or $z_{ra}$ to two rendering generators, ND changes from 0.52 of w/o the PRA's progressive way  to w/o $g_s$'s 0.37, i.e.,  the performance drops by  $29 \%$ compared to the full model without PRA's progressive way, since two generators lack the link to push themselves to use $z_s$ to model shared underlying structure.  It indicates the common network, i.e., underlying structure generator $g_s$, is essential to our model. In a word, both two elements of PRA ( i.e., PRA itself) are  essential to our model.

\textbf{Effect of shared parameters of discriminators and auxiliary losses}. When we remove partially shared parameters of two discriminators, ND drops by $4.6\%$, indicating weight-sharing constraint is helpful to improve our model's performance. We remove the auxiliary losses, i.e., the loss an loss
\cite{infoGAN} for reconstructing priors in Eqn. (\ref{eqn:lip}) and the loss for reconstructing real images in Eqn. (\ref{eqn:lrec}). We can see from Table \ref{tbl:ablation} that without the auxiliary losses, the disentanglability drops by $3.0\%$, illustrating that they have mild effect on regularizing our architecture to further improve model's performance.

As analyzed above, our proposed auxiliary domain and Progressive Rendering Architecture (PRA) do play a critical role in explicitly learning disentangled representation. 




\section{Conclusion}

In this work, we formulate the problem of image generation for explicitly disentangling underlying spatial structure and rendering and our proposed DSRGAN successfully learns disentangled presentation by introducing an auxiliary domain and designing a Progressive Rendering Architecture (PRA) in our framework.  Further, to evaluate our model, we propose the Normalized Disentanglability, which can reflect disentanglability of a generator well as shown in experiments. In Section \ref{sec:experiments}, we show that our model can effectively disentangle the underlying spatial structure and rendering in target domain and ablation study proves that the auxiliary domain and PRA are critical to our model.

{\small
\bibliographystyle{ieee}
\bibliography{egbib}
}

\end{document}